\documentclass[journal]{IEEEtran}
\usepackage{amsmath}
\usepackage{algorithm}
\usepackage{algorithmic}
\usepackage{graphicx}
\usepackage{multirow}
\usepackage{tabularx}
 
\usepackage{soul}
\usepackage{colortbl}  
\usepackage{color,xcolor}
\usepackage{array}
\usepackage{amssymb}
\usepackage{bbding}
\usepackage{subfigure}
\usepackage{diagbox}
\usepackage{rotating}
\usepackage{bm}
\usepackage{tikz} 
\usepackage{booktabs}
\usepackage{cite}
\usepackage{CJK}
\usepackage{overpic}
\usepackage[colorlinks,
            linkcolor=red,
            anchorcolor=blue,
            citecolor=black
            ]{hyperref}

\newcommand{\addFig}[1]{}
\newcommand{\addFigs}[1]{}

\usepackage{subfigure}
\newcommand{\etal}{\textit{et~al}.~}
\newcommand{\ie}{\textit{i}.\textit{e}.,~}

\definecolor{lightgreen}{HTML}{9aff99}
\definecolor{lightpink}{HTML}{FFE2E1}
\definecolor{lightblue}{HTML}{c9e5ff}
\definecolor{lightpurple}{HTML}{abacf4}
\definecolor{lightorange}{HTML}{ffcc67}
\definecolor{lightgray}{HTML}{c0c0c0}
\definecolor{lightred}{HTML}{FF8181}

\soulregister{\cite}7 
\soulregister{\citep}7 
\soulregister{\citet}7 
\soulregister{\ref}7 
\soulregister{\pageref}7 

\usepackage{balance}
\usepackage{cleveref}
\crefformat{section}{\S#2#1#3} 
\crefformat{subsection}{\S#2#1#3}
\crefformat{subsubsection}{\S#2#1#3}

\ifCLASSINFOpdf
\else
\fi

\begin{document}
%
\title{Salient Object Detection in Optical Remote Sensing Images Driven by Transformer}
%
%
%

\author{Gongyang~Li,
	Zhen~Bai,
	Zhi~Liu,~\IEEEmembership{Senior Member,~IEEE},
	Xinpeng~Zhang,~\IEEEmembership{Member,~IEEE},
	and~Haibin~Ling,~\IEEEmembership{Fellow,~IEEE}

\thanks{Gongyang Li, Zhen Bai, Zhi Liu, and Xinpeng Zhang are with Key Laboratory of Specialty Fiber Optics and Optical Access Networks, Joint International Research Laboratory of Specialty Fiber Optics and Advanced Communication, Shanghai Institute for Advanced Communication and Data Science, Shanghai University, Shanghai 200444, China, and School of Communication and Information Engineering, Shanghai University, Shanghai 200444, China. Gongyang Li and Zhi Liu are also with Wenzhou Institute of Shanghai University, Wenzhou 325000, China (email: ligongyang@shu.edu.cn; bz536476@163.com; liuzhisjtu@163.com; xzhang@shu.edu.cn).}
\thanks{Haibin Ling is with the Department of Computer Science, Stony Brook University, Stony Brook, NY 11794 USA (email: hling@cs.stonybrook.edu).}
\thanks{\textit{Corresponding authors: Zhi Liu and Xinpeng Zhang.}}
}

\markboth{IEEE TRANSACTIONS ON IMAGE PROCESSING}%
{Shell \MakeLowercase{\textit{et al.}}: Bare Demo of IEEEtran.cls for IEEE Journals}

\maketitle

\begin{abstract}
Existing methods for \emph{Salient Object Detection in Optical Remote Sensing Images} (ORSI-SOD) mainly adopt \emph{Convolutional Neural Networks} (CNNs) as the backbone, such as VGG and ResNet.
Since CNNs can only extract features within certain receptive fields, most ORSI-SOD methods generally follow the local-to-contextual paradigm.
In this paper, we propose a novel \emph{Global Extraction Local Exploration Network} (GeleNet) for ORSI-SOD following the global-to-local paradigm.
Specifically, GeleNet first adopts a transformer backbone to generate four-level feature embeddings with global long-range dependencies.
Then, GeleNet employs a \emph{Direction-aware Shuffle Weighted Spatial Attention Module} (D-SWSAM) and its simplified version (SWSAM) to enhance local interactions, and a \emph{Knowledge Transfer Module} (KTM) to further enhance cross-level contextual interactions.
D-SWSAM comprehensively perceives the orientation information in the lowest-level features through directional convolutions to adapt to various orientations of salient objects in ORSIs, and effectively enhances the details of salient objects with an improved attention mechanism.
SWSAM discards the direction-aware part of D-SWSAM to focus on localizing salient objects in the highest-level features.
KTM models the contextual correlation knowledge of two middle-level features of different scales based on the self-attention mechanism, and transfers the knowledge to the raw features to generate more discriminative features.
Finally, a saliency predictor is used to generate the saliency map based on the outputs of the above three modules.
Extensive experiments on three public datasets demonstrate that the proposed GeleNet outperforms relevant state-of-the-art methods.
The code and results of our method are available at https://github.com/MathLee/GeleNet.
\end{abstract}


\IEEEpeerreviewmaketitle

\section{Introduction}
\IEEEPARstart{S}{alient} Object Detection (SOD) focuses on finding and locating the most visually prominent objects/regions in a scene~\cite{19CRMCO,2015SODBenchmark,2020ICNet}.
It is a common pre-processing step for many tasks in computer vision, such as quality assessment~\cite{16SODIQA,19SGDNet}, object segmentation~\cite{LGY2019,LGY2021PFOS,2021seg,2021seg1,2023LASNet}, video compression~\cite{2014Compression}, and object tracking~\cite{2021Tracking}.
Recently, SOD in \emph{Optical Remote Sensing Images} (ORSI-SOD)~\cite{2023SeaNet,2021DAFNet,2019LVNet}, as an emerging topic, has attracted the attention of many researchers, and has been widely used in agriculture, forestry, environmental science, and security surveillance.

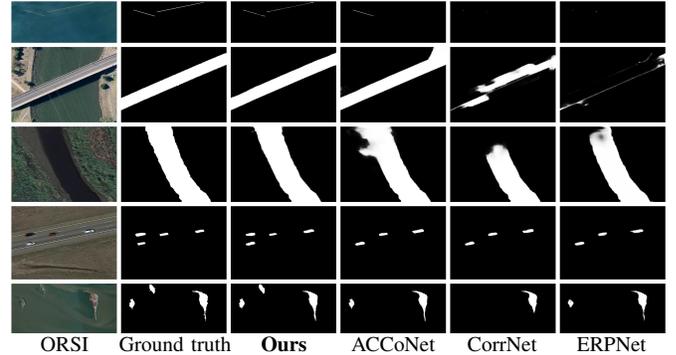
\begin{figure}[t!]
  \centering
  \footnotesize
  \begin{overpic}[width=1\columnwidth]{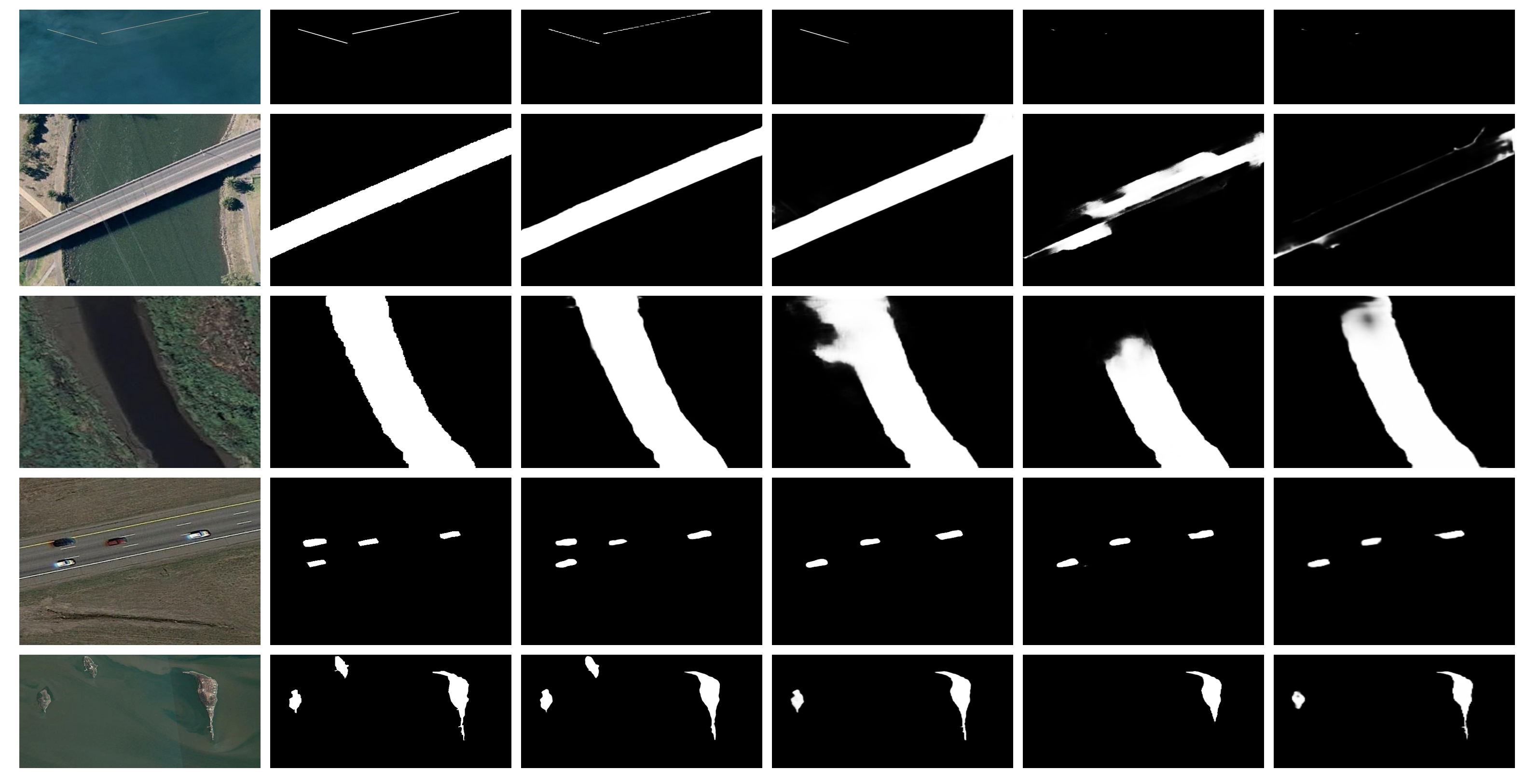}
      \put(4.4,-1.7){ ORSI}
      \put(16.2,-1.7){ Ground truth }   
      \put(37.5,-1.7){ \textbf{Ours}}
      \put(51.05,-1.7){ ACCoNet}      
      \put(68.5,-1.7){ CorrNet}  
      \put(84.9,-1.7){ ERPNet}
  \end{overpic}
  \caption{Saliency maps produced by our method and three state-of-the-art ORSI-SOD methods, including ACCoNet~\cite{2022ACCoNet}, CorrNet~\cite{2022CorrNet}, and ERPNet~\cite{2022ERPNet}.
  Please zoom in for details, especially the first row.
    }\label{fig:example}
\end{figure}
With the rapid development of deep learning, \emph{Convolutional Neural Networks} (CNNs)~\cite{1989CNN} have dominated the field of computer vision with their powerful feature representation capabilities.
Many effective CNN-based solutions for ORSI-SOD are proposed~\cite{2019LVNet,2022ACCoNet,2022CorrNet,2022MCCNet,2021SARNet,2022MJRBM,2022EMFINet,2022ERPNet}.
While a few methods follow the local-to-local paradigm~\cite{2022MCCNet,2022ERPNet}, most methods adopt the local-to-contextual paradigm\footnote{Here, the first ``local" in both paradigms specifically refers to using CNN backbones to extract features with limited receptive fields.}~\cite{2019LVNet,2022ACCoNet,2022CorrNet,2020PDFNet,2022RRNet,2022MJRBM,2022EMFINet,2021SARNet}.
Both paradigms first use a CNN backbone, such as VGG~\cite{2015VGG} and ResNet~\cite{2016ResNet}, to extract basic feature embeddings.
The local-to-local paradigm focuses on exploring valuable information in single-level feature embeddings.
Differently, since the local-to-contextual one considers that CNNs only extract features within certain receptive fields, it focuses on designing specific modules to mine the contextual information between feature embeddings at multiple levels.
The above paradigms promote the development of ORSI-SOD and achieve promising performance.

However, due to the characteristics of ORSI scenes, such as variation in object orientation, scale, and category, the above paradigms suffer from obvious limitations.
The local-to-local paradigm ignores contextual information that is useful for handling the above scenes.
The contextual information captured by the local-to-contextual paradigm is still based on convolution layers with limited receptive fields, which is also insufficient to handle challenging scenes of ORSIs.
For intuitive understanding, we show the saliency maps generated by typical methods for both paradigms in Fig.~\ref{fig:example}, where ACCoNet~\cite{2022ACCoNet} and CorrNet~\cite{2022CorrNet} belong to the local-to-contextual paradigm, and ERPNet~\cite{2022ERPNet} belongs to the local-to-local paradigm. 
We find that these methods suffer from orientation insensitivity, incomplete detection, and missing salient objects.

Inspired by the above observations, in this paper, we propose to design new solutions following the global-to-local paradigm.
Our main idea is to replace the CNN backbone with a transformer one that can establish global relationships (\ie changing the first ``local" in two existing paradigms to ``global") and to perform local enhancement on the extracted global features.
With this idea, we build a novel \emph{Global Extraction Local Exploration Network} (GeleNet) for ORSI-SOD with a transformer backbone.
Transformers~\cite{2017transformer,2021ViT,2021T2T,2021Swin} are known to be good at modeling the global long-range dependencies between feature patches. This unique ability of transformer enables GeleNet to deal with complex scenes and changeable objects in ORSIs.
Furthermore, in GeleNet, we focus on local and cross-level contextual interactions, which are beneficial for highlighting salient objects in ORSIs.

In particular, we adopt the popular PVT~\cite{2021PVT,2022PVTv2} as the backbone of our GeleNet.
To alleviate the orientation insensitivity issue of previous methods, we propose a \emph{Direction-aware Shuffle Weighted Spatial Attention Module} (D-SWSAM), and assign it to the lowest-level features to adequately identify the orientation of objects through directional convolutions with four directions.
D-SWSAM is also equipped with an improved attention mechanism to outline the details of salient objects.
Since high-level features contain location information rather than orientation and texture information, we extract the corresponding part containing the improved attention mechanism from D-SWSAM, \ie SWSAM, and assign it to the highest-level features to determine the location of salient objects.
The above modules can well enhance local interactions of intra-level features.
In addition, we propose a \emph{Knowledge Transfer Module} (KTM) for the remaining adjacent features to explore contextual interactions between inter-level features and transfer the specific knowledge of salient objects between adjacent features to the raw features.
In this way, the proposed GeleNet can generate saliency maps with accurate orientations and complete objects, as illustrated in the third column of Fig.~\ref{fig:example}, and consistently outperforms compared methods on three datasets.

Our main contributions are summarized in three aspects:
\begin{itemize}
\item We propose a transformer-based ORSI-SOD solution, \emph{GeleNet}, with the global-to-local paradigm, which is different from the local-to-contextual paradigm followed by most existing CNN-based methods.
To the best of our knowledge, this is the first transformer-driven ORSI-SOD solution.

\item We propose the D-SWSAM and its variant SWSAM to enhance local interactions of the extracted global feature embeddings.
D-SWSAM can tackle the problem of objects with various orientations in ORSIs and enhance the details of salient objects in the lowest-level features, while SWSAM can locate salient objects in the highest-level features.

\item We propose the KTM to enhance contextual interactions of two middle-level features.
In KTM, we model the contextual correlation knowledge of two types of combinations (\ie product and sum) of these features, and transfer the knowledge to the raw features to generate more discriminative features.

\end{itemize}

The rest of this paper is arranged as follows. 
In Sec.~\ref{sec:related}, we review the related work.
In Sec.~\ref{sec:OurMethod}, we describe the details of the proposed GeleNet.
In Sec.~\ref{sec:exp}, we conduct comprehensive experiments and ablation studies.
In Sec.~\ref{sec:con}, we present the conclusion.

\section{Related Work}
\label{sec:related}

\subsection{Salient Object Detection in Optical Remote Sensing Images}
\label{sec:ORSI_SOD}
Salient object detection in optical remote sensing images plays an important role in understanding ORSIs.
Recently, with the successive construction of the three datasets~\cite{2019LVNet,2021DAFNet,2022MJRBM}, numerous ORSI-SOD methods are proposed.
Here we focus on CNN-based methods, which dominate this topic and achieve promising performance.

Existing CNN-based ORSI-SOD methods mainly follow two paradigms, \ie the local-to-local paradigm and the local-to-contextual paradigm.
The local-to-local paradigm typically extracts feature embeddings containing local information through the CNN backbone, and then explores valuable information in single-level feature embeddings.
For example, in~\cite{2022ERPNet}, Zhou~\etal extracted multi-level features through the CNN backbone, and performed edge extraction and feature fusion on each level of features in two parallel decoders.
Li~\etal\cite{2022MCCNet} explored the complementarity of foreground, edge, background, and the global image-level content of single-level features, and aimed at generating complete salient objects.
They focused on the extraction of various specific information on single-level features (\ie local features), ignoring the contextual interactions between local features at different levels.

The local-to-contextual paradigm, by contrast, explores contextual information between local feature embeddings at different levels, and is therefore popularly adopted by recent solutions.
For example, Li~\etal\cite{2019LVNet} extracted multi-level features from multiple inputs, and employed nested connections to aggregate them.
Similarly, Zhou~\etal\cite{2022EMFINet} proposed a cascaded feature fusion module to fuse multi-level features from different branches.
In~\cite{2021SARNet}, Huang~\etal aggregated three high-level features to produce contextual semantic information to approximately locate salient objects.
Li~\etal\cite{2022CorrNet} proposed a correlation module for continuous semantic features, generating an initial coarse saliency map for location guidance of low-level features.
Tu~\etal\cite{2022MJRBM} proposed two decoders to aggregate three adjacent features twice with salient boundary features.
Li~\etal\cite{2022ACCoNet} designed a specific module for adjacent features, aiming at coordinating cross-scale interactions and mining valuable contextual information.

Despite great progress achieved by the local-to-contextual paradigm, the explored contextual interactions only mine interactions between features at different levels through convolution-based modules.
In this paper, inspired by the popular transformer~\cite{2017transformer,2021ViT,2021T2T,2021Swin,2021PVT}, we propose the global-to-local paradigm that first models the global long-range dependencies between feature patches and then enhances the local and contextual interactions, and build a novel GeleNet for ORSI-SOD.
Benefiting from the global view of the transformer and the local enhancement of our proposed modules, our GeleNet can better perceive salient objects with numerous scales, diverse types, and multiple numbers in ORSIs.

\subsection{Salient Object Detection with Transformer}
\label{sec:	SODwithViT}
Transformer was first proposed for \emph{Natural Language Processing} (NLP)~\cite{2017transformer}, which is good at modeling global long-range dependencies between word vectors.
Following its success in NLP, researchers have extended it into computer vision and achieved remarkable progress on numerous tasks, especially on dense prediction tasks~\cite{2021PVT,2022PVTv2,2021Swin}.

Here, we introduce some representative works on transformer-based SOD, involving SOD in \emph{Natural Scene Images} (NSI-SOD)~\cite{2021TFcod,2021VST}, RGB-D/T SOD~\cite{2021SwinNet,2021TriTransNet,2022grouptf}, co-saliency detection~\cite{2022unified}, and video SOD~\cite{2022unified}.
In general, transformer-based SOD methods can be roughly divided into three types depending on where the transformer is used.
The first type of method adopted transformer as the feature extractor in the encoding phase.
For instance, Liu~\etal\cite{2021SwinNet} used Swin Transformer~\cite{2021Swin} to extract basic features from RGB-D/T pairs, and aligned cross-modality features through attention mechanism to generate discriminative features.
Liu~\etal\cite{2021TFcod} achieved effective context modeling using the same backbone as~\cite{2021SwinNet} for NSI-SOD.
The second type of method adopted transformer to develop modules in the decoding phase.
Liu~\etal\cite{2021TriTransNet} proposed a triplet transformer embedding module to enhance high-level features by learning long-range dependencies across layers.
In~\cite{2022unified}, Su~\etal proposed a unified transformer framework for co-saliency detection and video SOD, which is equipped with two transformer blocks to capture the long-range dependencies among a group of features from different images/frames.
Fang~\etal\cite{2022grouptf} proposed a multiple transformer module to learn the common information of cross-modality and cross-scale features.
The last type of method utilized the pure transformer architecture to achieve SOD.
Liu~\etal\cite{2021VST} adopted T2T-ViT~\cite{2021T2T} as the backbone, and proposed a multi-task transformer decoder to jointly detect salient objects and boundaries.

The above transformer-based SOD methods achieve impressive results on specific SOD tasks.
Therefore, we introduce the transformer into the ORSI-SOD task, and propose the first transformer-driven ORSI-SOD method, \ie GeleNet.
Our method belongs to the first type of method, and adopts PVT~\cite{2021PVT,2022PVTv2} as the backbone to extract long-range dependency features from input ORSIs.

\subsection{Attention Mechanism}
\label{sec:Attention}
Attention mechanism is widely used in computer vision and image analysis.
In general, it includes channel attention~\cite{2020SENet}, spatial attention~\cite{2018CBAM}, and self-attention~\cite{2017transformer,2019DANet}.
SENet~\cite{2020SENet} was a classic channel attention model, which explicitly represents dependencies between channels to adaptively recalibrate features.
ECANet~\cite{2020ECANet} developed an extremely lightweight channel attention module through a fast 1D convolution.
Moreover, CBAM~\cite{2018CBAM} additionally introduced spatial attention, and inferred attention maps along channel and spatial domain in turn for adaptive feature enhancement.
Li~\etal\cite{2019SGE} proposed the \emph{Spatial Group-wise Enhance} (SGE), which first splits features into several sub-features, then extracts specific semantics from each sub-feature, and finally adjusts the importance of semantics of each sub-feature by an attention factor.
Zhang~\etal\cite{2021SANet} proposed a lightweight shuffle attention, which also first splits features into several groups, then performs channel attention and spatial attention in parallel, and finally introduces channel shuffle to allow information communication along channels.

Both SGE~\cite{2019SGE} and shuffle attention~\cite{2021SANet} consider only the attention of each sub-feature, but ignore the consistency of attention between different sub-features, which is not friendly to SOD.
In addition, since the global features extracted by the transformer lack channel interaction, it is unreasonable for shuffle attention to put the shuffle operation at the end.
Therefore, we propose an improved spatial attention module, namely SWSAM, which focuses on enhancing the channel interactions of global features and improving the effectiveness of spatial attention to highlight salient regions more accurately.
Notably, we further integrate SWSAM and directional convolutions, and propose D-SWSAM to adapt to various orientations of salient objects in ORSIs.
Moreover, we also propose a self-attention-based KTM to model and transfer the contextual knowledge to generate more discriminative features.

\section{Proposed Method}
\label{sec:OurMethod}
In this section, we elaborate on the proposed transformer-driven GeleNet.
In Sec.~\ref{sec:Overview}, we depict the network overview.
In Sec.~\ref{sec:D-SWSAM} and Sec.~\ref{sec:KTM}, we introduce D-SWSAM and KTM, respectively.
In Sec.~\ref{sec:Saliency_Predictor}, we present the saliency predictor and loss function.

\begin{figure*}
	\centering
	\begin{overpic}[width=1\textwidth]{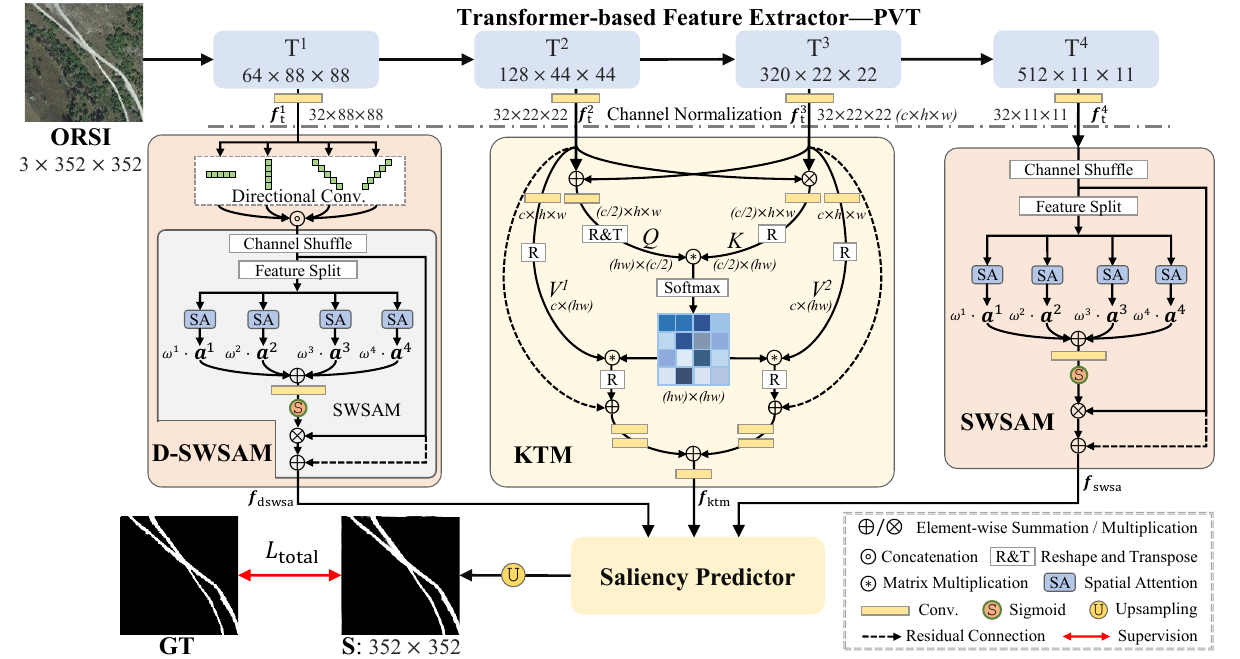}
    \end{overpic}
	\caption{Pipeline of the proposed transformer-driven GeleNet, which consists of a feature extractor, three modules, and a saliency predictor.
	First, we adopt a transformer-based feature extractor PVT-v2-b2~\cite{2022PVTv2} to extract four-level basic feature embeddings with global long-range dependencies.
	Then, we employ the \emph{Direction-aware Shuffle Weighted Spatial Attention Module} (D-SWSAM), the \emph{Knowledge Transfer Module}, and the variant of D-SWSAM (\ie SWSAM) to deal with the corresponding features, respectively.
	Specifically, in D-SWSAM, we perform four directional convolutions with different directions (\ie horizontal, vertical, leading diagonal, and reverse diagonal) on the lowest-level features to extract specific orientation information, and then use SWSAM to outline the details regions.
	We also adopt SWSAM to enhance the location of salient objects in the highest-level features.
	In KTM, we model the contextual correlation knowledge of two types of combinations (\ie product and sum) of two middle-level features, and transfer the knowledge to the raw features to generate more discriminative features.
	Finally, we use a saliency predictor to generate a saliency map from the outputs of the above modules.
    }
    \label{fig:Framework}
\end{figure*}


\subsection{Network Overview}
\label{sec:Overview}
As illustrated in Fig.~\ref{fig:Framework}, the proposed GeleNet follows the popular three-stage structure~\cite{20CMWNet,2021AGCNet} in SOD, including a feature extractor for basic feature generation, three modules (\ie D-SWSAM, KTM, and SWSAM) for feature interaction/enhancement, and a saliency predictor for saliency map generation.

Concretely, we use the \emph{Pyramid Vision Transformer} (PVT)~\cite{2022PVTv2} as the backbone, whose input size is set to $3\!\times\!352\!\times\!352$.
PVT consists of four transformer encoder blocks denoted as T$^{i}$ ($i \in \{1,2,3,4\}$), and can generate four-level basic global features denoted as $\boldsymbol{\hat{f}}^{i}_{\rm t} \in \mathbb{R}^{c_i\!\times\!h_i\!\times\!w_i}$, where $c_i \in \{64,128,320,512\}$, and $h_i/w_i=\frac{352}{2^{i+1}}$.
To improve the computational efficiency, we unify the channel number of $\boldsymbol{\hat{f}}^{i}_{\rm t}$ ($i \in \{1,3,4\}$) to 32 by the channel normalization (\ie a convolution layer), generating $\boldsymbol{f}^{i}_{\rm t} \in \mathbb{R}^{c\!\times\!h_i\!\times\!w_i}$, where $c$ is 32.
Notably, for $\boldsymbol{\hat{f}}^{2}_{\rm t}$, we not only reduce its channel number to 32, but also adjust its resolution from 44$\times$44 to 22$\times$22 for subsequent processing in KTM, generating $\boldsymbol{f}^{2}_{\rm t} \in \mathbb{R}^{32\!\times\!22\!\times\!22}$.
For the lowest-level features $\boldsymbol{f}^{1}_{\rm t}$ and the highest-level features $\boldsymbol{f}^{4}_{\rm t}$, we adopt an improved spatial attention mechanism for local enhancement.
According to the characteristics of features at different levels, we adopt D-SWSAM for $\boldsymbol{f}^{1}_{\rm t}$ to extract orientation information and achieve local detail enhancement, generating $\boldsymbol{{f}}_{\rm dswsa}$.
While we adopt SWSAM for $\boldsymbol{f}^{4}_{\rm t}$ to achieve local location enhancement, generating $\boldsymbol{{f}}_{\rm swsa}$.
Moreover, we adopt KTM to activate cross-level contextual interactions of $\boldsymbol{f}^{2}_{\rm t}$ and $\boldsymbol{f}^{3}_{\rm t}$, generating discriminative features $\boldsymbol{{f}}_{\rm ktm}$.
Taking advantage of PVT and these three novel modules, we infer salient objects in the saliency predictor, which is a variant of the effective partial decoder~\cite{2019CPD}.

\subsection{Direction-aware Shuffle Weighted Spatial Attention Module}
\label{sec:D-SWSAM}
Since the basic features extracted by PVT is with global long-range dependencies, we want to explore their local enhancements to complement their global information and adapt to complex scenes in ORSIs.
To be precise, we hope to consistently highlight salient regions in features across different channels, which is important for SOD.
Traditional spatial attention~\cite{2018CBAM} is known to be effective way to achieve this goal, however, it generates the spatial attention map in a global manner. Specifically, it performs global max pooling and global average pooling on all channels, which may produce an insufficient spatial attention map.
Differently, SGE~\cite{2019SGE}, as a grouping attention, splits features into several subsets and generates a specific spatial attention map from each sub-feature for individual enhancement.
While considering only the attention of each sub-feature, SGE ignores the consistency of attention between different sub-features, resulting in the lack of consistency in the group-enhanced features, which is not friendly to SOD.
Inspired by\cite{2018CBAM,2019SGE}, we propose an effective grouping spatial attention mechanism for SOD, \ie the \emph{Shuffle Weighted Spatial Attention Module} (SWSAM), which first generates the local spatial attention map from each sub-feature, and then adopts the weighted fusion operation to produce the final spatial attention map for consistent enhancement.

In addition, salient objects in ORSIs usually have various orientations, as shown in Fig.~\ref{fig:example} and Fig.~\ref{fig:Framework}, which often bring troubles to existing methods using the traditional convolutions.
To solve this issue, we specifically introduce directional convolutions with different directions~\cite{2021DirConv} into SWSAM, and propose D-SWSAM to explicitly extract orientation information of salient objects and achieve local enhancement.
Moreover, we arrange D-SWSAM to deal with $\boldsymbol{f}^{1}_{\rm t}$.
The detailed structure of D-SWSAM is presented in the left part of Fig.~\ref{fig:Framework}.
In the following, we elaborate D-SWSAM in three parts, \ie the directional convolution unit, the channel shuffle and feature split, and the weighted spatial attention, of which the latter two parts constitute SWSAM.

\textit{1) Directional Convolution Unit}.
The directional convolution unit takes into account the four basic directions, and is composed of four directional convolution layers with horizontal (h), vertical (v), leading diagonal (ld), and reverse diagonal (rd) directions~\cite{2021DirConv}.
We parallelize these four directional convolution layers to simultaneously mine different orientation information of $\boldsymbol{f}^{1}_{\rm t}$, and concatenate the results for information integration.
We formulate the above process as follows: 
\begin{equation}
   \begin{aligned}
    \boldsymbol{f}_{\rm ori} = {\rm conv_{h}}(\boldsymbol{f}^{1}_{\rm t}) \circledcirc {\rm conv_{v}}(\boldsymbol{f}^{1}_{\rm t}) \circledcirc {\rm conv_{ld}}(\boldsymbol{f}^{1}_{\rm t}) \circledcirc {\rm conv_{rd}}(\boldsymbol{f}^{1}_{\rm t}),
    \label{eq:1}
    \end{aligned}
\end{equation}
where $\boldsymbol{f}_{\rm ori} \in \mathbb{R}^{32\!\times\!88\!\times\!88}$ denotes the output orientation features, $ \circledcirc$ is the concatenation, ${\rm conv_{x}}(\cdot)$ is the directional convolution layer with the direction ${\rm x \in \{h,v,ld,rd\}}$.
Considering the input feature size and computational efficiency, here we set the kernel size and the output channel of four directional convolutions to 5 and 8, respectively.
To show the extracted orientation information intuitively, we expand $\boldsymbol{f}_{\rm ori}$ across channel as $ [\boldsymbol{f}^{1}_{\rm h}, ..., \boldsymbol{f}^{8}_{\rm h}, \boldsymbol{f}^{1}_{\rm v}, ..., \boldsymbol{f}^{8}_{\rm v}, \boldsymbol{f}^{1}_{\rm ld}, ..., \boldsymbol{f}^{8}_{\rm ld}, \boldsymbol{f}^{1}_{\rm rd}, ..., \boldsymbol{f}^{8}_{\rm rd}]$, where $\boldsymbol{f}_{\rm x} \in \mathbb{R}^{1\!\times\!88\!\times\!88}$ is a single-channel feature and we omit its superscript, and each directional convolution layer generates an eight-channel feature.

\textit{2) Channel Shuffle and Feature Split}.
Inspired by ShuffleNet~\cite{2017ShuffleNet} and shuffle attention~\cite{2021SANet}, which shuffle features to achieve information communication along channels, we shuffle $\boldsymbol{f}_{\rm ori}$ with four groups to evenly disperse the orientation information, achieving $\boldsymbol{f}_{\rm shuf} \in \mathbb{R}^{32\!\times\!88\!\times\!88}$, which can be expanded as
$[\boldsymbol{f}^{1}_{\rm h}, \boldsymbol{f}^{1}_{\rm v}, \boldsymbol{f}^{1}_{\rm ld}, \boldsymbol{f}^{1}_{\rm rd}, ..., \boldsymbol{f}^{4}_{\rm h}, \boldsymbol{f}^{4}_{\rm v}, \boldsymbol{f}^{4}_{\rm ld}, \boldsymbol{f}^{4}_{\rm rd}, ..., \boldsymbol{f}^{8}_{\rm h}, \boldsymbol{f}^{8}_{\rm v}, \boldsymbol{f}^{8}_{\rm ld}, \boldsymbol{f}^{8}_{\rm rd}]$.

Then, we split $\boldsymbol{f}_{\rm shuf}$ into four feature subsets along channel, generating
$\{\boldsymbol{f}^{1}_{\rm s-shuf}, \boldsymbol{f}^{2}_{\rm s-shuf}, \boldsymbol{f}^{3}_{\rm s-shuf}, \boldsymbol{f}^{4}_{\rm s-shuf} \} \in \mathbb{R}^{8\!\times\!88\!\times\!88}$, where $\boldsymbol{f}^{n}_{\rm s-shuf}$ ($n \in \{1,2,3,4\}$) can be expanded as
$[\boldsymbol{f}^{2n-1}_{\rm h}, \boldsymbol{f}^{2n-1}_{\rm v}, \boldsymbol{f}^{2n-1}_{\rm ld}, \boldsymbol{f}^{2n-1}_{\rm rd},\boldsymbol{f}^{2n}_{\rm h}, \boldsymbol{f}^{2n}_{\rm v}, \boldsymbol{f}^{2n}_{\rm ld}, \boldsymbol{f}^{2n}_{\rm rd}]$. 
The above operations activate the interaction between features of different orientations, so that each sub-feature evenly contains orientation information in four directions, which is conducive to generating an accurate spatial attention map for each sub-feature.

\textit{3) Weighted Spatial Attention}.
We then apply the traditional spatial attention~\cite{2018CBAM} to the above sub-features $\boldsymbol{f}^{n}_{\rm s-shuf}$, generating corresponding spatial attention maps $\boldsymbol{a}^{n} \in (0,1)^{1\!\times\!88\!\times\!88}$ as follows:
\begin{equation}
   \begin{aligned}
     \boldsymbol{a}^{n} =  {\rm SA}(\boldsymbol{f}^{n}_{\rm s-shuf}),
    \label{eq:2}
    \end{aligned}
\end{equation}
where ${\rm SA}(\cdot)$ is the spatial attention operation.
These four spatial attention maps can extract salient regions in local sub-features comprehensively without neglecting salient regions in the original complete $\boldsymbol{f}_{\rm ori}$.

Next, we design a learnable attention fusion approach, that is, set a learnable parameter ${w}^{n} \in [0,1]$ for each spatial attention map $\boldsymbol{a}^{n}$ and aggregate them as follows:
\begin{equation}
   \begin{aligned}
     \boldsymbol{a}_{\rm ori} =  {\rm sigmoid}( {\rm conv}( \sum_{n=1}^4 {w}^{n} \cdot \boldsymbol{a}^{n} ) ),
    \label{eq:3}
    \end{aligned}
\end{equation}
where $\boldsymbol{a}_{\rm ori} \in (0,1)^{1\!\times\!88\!\times\!88}$ is the aggregated spatial attention map, ${w}^{n} $ is initialized as 0.25 and gradually converges to appropriate weights, $\sum_{n=1}^4 {w}^{n} =1$, ${\rm conv}(\cdot)$ is the normal convolution layer, and ${\rm sigmoid}(\cdot)$ is the sigmoid activation function.
In this way, we can obtain a comprehensive and orientation-sensitive spatial attention map $\boldsymbol{a}_{\rm ori}$.
We adopt $\boldsymbol{a}_{\rm ori}$ to achieve consistent detail enhancement, generating the output feature of D-SWSAM $ \boldsymbol{f}_{\rm dswsa} \in \mathbb{R}^{32\!\times\!88\!\times\!88}$ as follows:
\begin{equation}
   \begin{aligned}
     \boldsymbol{f}_{\rm dswsa} =  (\boldsymbol{a}_{\rm ori} \otimes \boldsymbol{f}_{\rm shuf} ) \oplus \boldsymbol{f}_{\rm shuf} ,
    \label{eq:4}
    \end{aligned}
\end{equation}
where $\otimes$ is the element-wise multiplication and $\oplus$ is the element-wise summation.
Notably, here we perform detail enhancement on $\boldsymbol{f}_{\rm shuf}$ rather than $\boldsymbol{f}_{\rm ori}$, which continues to maintain valid channel interactions.

\textit{4) Applying SWSAM for Location Enhancement}.
As shown in Fig.~\ref{fig:Framework}, instead of D-SWSAM, we apply SWSAM on the highest-level features $\boldsymbol{f}^{4}_{\rm t}$ for location enhancement.
This is because $\boldsymbol{f}^{4}_{\rm t}$ mainly contains location information, rather than detail information such as orientation information and texture information, which means that the directional convolution unit in D-SWSAM is superfluous.
Therefore, we abandon this unit.
In addition, $\boldsymbol{f}^{4}_{\rm t}$ is extracted using PVT which focuses on modeling the long-range dependencies between feature patches and inevitably ignores feature interactions between channels.
So we maintain the channel shuffle operation in SWSAM to explicitly increase the channel interaction.
In this way, we can obtain the output feature of SWSAM $ \boldsymbol{f}_{\rm swsa} \in \mathbb{R}^{32\!\times\!11\!\times\!11}$.

In summary, our D-SWSAM and SWSAM are designed according to specific characteristics of extracted global features of ORSIs to better enhance local interactions.
We believe our D-SWSAM can effectively assist GeleNet to adapt to salient objects with various orientations in ORSIs, and our SWSAM can assist GeleNet to accurately locate all salient objects in ORSIs.

\subsection{Knowledge Transfer Module}
\label{sec:KTM} 
For the lowest-level and highest-level features, we design special modules to process them to achieve local interactions according to their respective characteristics.
However, it is insufficient to consider only local enhancement, we enhance cross-level contextual interactions on two middle-level features (\ie $\boldsymbol{f}^{2}_{\rm t}$ and $\boldsymbol{f}^{3}_{\rm t}$) to explore the discriminative information of salient objects.
Inspired by the self-attention mechanism~\cite{2017transformer,2019DANet}, we propose a knowledge transfer module to achieve the goal.
The detailed structure of KTM is presented in the middle part of Fig.~\ref{fig:Framework}.
In the following, we introduce the two KTM components, \ie the contextual correlation knowledge modeling and the knowledge transfer.

\textit{1) Contextual Correlation Knowledge Modeling}.
In SOD, the product of two features can reveal the significant information co-existing in both features, which is conducive to collaboratively identifying objects.
The sum of two features can comprehensively capture the information contained in both features without omission, which is conducive to elaborating objects.
In particular for our framework, the product and sum of $\boldsymbol{f}^{2}_{\rm t}$ and $\boldsymbol{f}^{3}_{\rm t}$ are complementary to a certain extent.
Therefore, we adopt self-attention~\cite{2017transformer,2019DANet} to model the contextual correlation knowledge between the product and sum of $\boldsymbol{f}^{2}_{\rm t}$ and $\boldsymbol{f}^{3}_{\rm t}$.

As stated in Sec.~\ref{sec:Overview}, we have unified the size of $\boldsymbol{f}^{2}_{\rm t}$ and $\boldsymbol{f}^{3}_{\rm t}$ to 32$\times$22$\times$22.
For convenience, we denote the size of $\boldsymbol{f}^{2}_{\rm t}$ and $\boldsymbol{f}^{3}_{\rm t}$ to $c\times\!h\times\!w$, as shown in Fig.~\ref{fig:Framework}.
Here, we denote the product and sum of $\boldsymbol{f}^{2}_{\rm t}$ and $\boldsymbol{f}^{3}_{\rm t}$ as $\boldsymbol{f}_{\rm pro} \in \mathbb{R}^{c\!\times\!h\!\times\!w} $ and $\boldsymbol{f}_{\rm sum} \in \mathbb{R}^{c\!\times\!h\!\times\!w}$, respectively.
As the KTM illustrated in Fig.~\ref{fig:Framework}, to reduce the computational cost, we perform a convolution layer with the channel number of $c/2$ on $\boldsymbol{f}_{\rm pro}$ and $\boldsymbol{f}_{\rm sum}$ to generate two new features $\{\boldsymbol{\tilde{f}}_{\rm pro},\boldsymbol{\tilde{f}}_{\rm sum} \}\in \mathbb{R}^{(c/2)\!\times\!h\!\times\!w}$.
Then, we reshape and transpose $\boldsymbol{\tilde{f}}_{\rm sum}$ to obtain $\boldsymbol{{f}}_{\rm Q} \in \mathbb{R}^{(hw)\!\times\!(c/2)} $, and reshape $\boldsymbol{\tilde{f}}_{\rm pro}$ to obtain $\boldsymbol{{f}}_{\rm K} \in \mathbb{R}^{(c/2)\!\times\!(hw)} $.
After that we model the contextual correlation knowledge $\mathrm{\mathbf{C}}\!\in\!\mathbb{R}^{(hw)\times\!(hw)}$ between $\boldsymbol{{f}}_{\rm Q}$ and $\boldsymbol{{f}}_{\rm K}$ as follows:
\begin{equation}
   \begin{aligned}
     \mathrm{\mathbf{C}} =  {\rm softmax}( \boldsymbol{{f}}_{\rm Q} \circledast \boldsymbol{{f}}_{\rm K} ),
    \label{eq:5}
    \end{aligned}
\end{equation}
where ${\rm softmax}(\cdot)$ is the softmax activation function and $\circledast$ is the matrix multiplication.
In this way, we model the pixel-to-pixel dependencies between the co-existing significant information of $\boldsymbol{f}_{\rm pro}$ and the comprehensive information of $\boldsymbol{f}_{\rm sum}$, which are effective to avoid missing salient regions/objects in ORSIs.

\textit{2) Knowledge Transfer}.
Meanwhile, we use a convolution layer on $\boldsymbol{f}^{2}_{\rm t}$ and $\boldsymbol{f}^{3}_{\rm t}$ to generate two new features $\{\boldsymbol{\tilde{f}}^{2}_{\rm t},\boldsymbol{\tilde{f}}^{3}_{\rm t} \}\in \mathbb{R}^{c\!\times\!h\!\times\!w}$, and then reshape them to obtain $\{\boldsymbol{{f}}_{\rm V^{1}}, \boldsymbol{{f}}_{\rm V^{2}} \}\in \mathbb{R}^{c\!\times\!(hw)}$.
After that we transfer the modeled knowledge $\mathrm{\mathbf{C}}$ to $\boldsymbol{{f}}_{\rm V^{1}}$ and $\boldsymbol{{f}}_{\rm V^{2}}$ to generate the informative transferred features $\{\boldsymbol{f}^{1}_{\rm tsf}, \boldsymbol{f}^{2}_{\rm tsf} \}\in \mathbb{R}^{c\!\times\!h\!\times\!w}$ as follows:
\begin{equation}
   \begin{aligned}
     \boldsymbol{f}^{1}_{\rm tsf} =  {\rm R}( \boldsymbol{{f}}_{\rm V^{1}} \circledast {\rm T}(\mathrm{\mathbf{C}}) ), \\
     \boldsymbol{f}^{2}_{\rm tsf} =  {\rm R}( \boldsymbol{{f}}_{\rm V^{2}} \circledast {\rm T}(\mathrm{\mathbf{C}}) ),
    \label{eq:6}
    \end{aligned}
\end{equation}
where ${\rm R}(\cdot)$ and ${\rm T}(\cdot)$ mean reshape and transpose, respectively.
Following~\cite{2019DANet}, we introduce a trainable weight to adaptively fuse $\boldsymbol{f}^{1}_{\rm tsf}$ and raw $\boldsymbol{f}^{2}_{\rm t}$ through residual connection, and do the same for $\boldsymbol{f}^{2}_{\rm tsf}$ and raw $\boldsymbol{f}^{3}_{\rm t}$, generating $\{\boldsymbol{\tilde{f}}^{1}_{\rm tsf}, \boldsymbol{\tilde{f}}^{2}_{\rm tsf} \}\in \mathbb{R}^{c\!\times\!h\!\times\!w}$.
Finally, we adopt an element-wise summation and a convolution layer to integrate the cross-level $\boldsymbol{\tilde{f}}^{1}_{\rm tsf}$ and $\boldsymbol{\tilde{f}}^{2}_{\rm tsf}$, generating the discriminative output feature of KTM $\boldsymbol{{f}}_{\rm ktm} \in \mathbb{R}^{c\!\times\!h\!\times\!w}$.

In summary, $\boldsymbol{{f}}_{\rm ktm}$ inherits the properties of two combinations of $\boldsymbol{f}^{2}_{\rm t}$ and $\boldsymbol{f}^{3}_{\rm t}$, so it has the ability to simultaneously identify and elaborate salient objects.
In addition, compared to $\boldsymbol{{f}}_{\rm dswsa}$ and $\boldsymbol{{f}}_{\rm swsa}$, $\boldsymbol{{f}}_{\rm ktm}$ is more contextual, which is beneficial for our GeleNet to combine with local enhanced features (\ie $\boldsymbol{{f}}_{\rm dswsa}$ and $\boldsymbol{{f}}_{\rm swsa}$) for better salient object inference.

\begin{table*}[t!]
  \centering
  \small
  \renewcommand{\arraystretch}{1.25}
  \renewcommand{\tabcolsep}{0.8mm}
  \caption{
   Quantitative comparisons with state-of-the-art NSI-SOD and ORSI-SOD methods on EORSSD and ORSSD datasets.
   $\downarrow$ indicates that the lower the better, while $\uparrow$ the opposite.
   We mark the top two results in \textcolor{red}{\textbf{red}} and \textcolor{blue}{\textbf{blue}}, respectively.
    }
\label{table:QuantitativeResults}
  
\resizebox{1\textwidth}{!}{
\begin{tabular}{r|c|cccccccc|cccccccc}
\midrule[1pt]    
 \multirow{2}{*}{\normalsize{Methods}}
 & \multirow{2}{*}{\normalsize{Type}}
 & \multicolumn{8}{c|}{EORSSD~\cite{2021DAFNet}} 
 & \multicolumn{8}{c}{ORSSD~\cite{2019LVNet}}  \\
 
 \cline{3-10} \cline{11-18} 
       &  & $S_{\alpha}\uparrow$ & $F_{\beta}^{\rm{max}}\uparrow$ & $F_{\beta}^{\rm{mean}}\uparrow$ & $F_{\beta}^{\rm{adp}}\uparrow$ & $E_{\xi}^{\rm{max}}\uparrow$ & $E_{\xi}^{\rm{mean}}\uparrow$ & $E_{\xi}^{\rm{adp}}\uparrow$ & $ \mathcal{M}\downarrow$
   	          & $S_{\alpha}\uparrow$ & $F_{\beta}^{\rm{max}}\uparrow$ & $F_{\beta}^{\rm{mean}}\uparrow$ & $F_{\beta}^{\rm{adp}}\uparrow$ & $E_{\xi}^{\rm{max}}\uparrow$ & $E_{\xi}^{\rm{mean}}\uparrow$ & $E_{\xi}^{\rm{adp}}\uparrow$ & $ \mathcal{M}\downarrow$\\
	     
\midrule[1pt]
R3Net$_{18}$~\cite{2018R3Net}   	& CN & .8184 & .7498 & .6302 & .4165 & .9483 & .8294 & .6462 & .0171
									 & .8141 & .7456 & .7383 & .7379 & .8913 & .8681 & .8887 &  .0399\\
PoolNet$_{19}$~\cite{2019PoolNet}  & CN & .8207 & .7545 & .6406 & .4611 & .9292 & .8193 & .6836 & .0210
									    & .8403 & .7706 & .6999 & .6166 & .9343 & .8650 & .8124 & .0358 \\ 
EGNet$_{19}$~\cite{2019EGNet}  	& CN & .8601 & .7880 & .6967 & .5379 & .9570 & .8775 & .7566 & .0110  
									 & .8721 & .8332 & .7500 & .6452 & .9731 & .9013 & .8226 & .0216 \\ 
GCPA$_{20}$~\cite{2020GCPA}  	& CN & .8869 & .8347 & .7905 & .6723 & .9524 & .9167 & .8647 & .0102  
									 & .9026 & .8687 & .8433 & .7861 & .9509 & .9341 & .9205 & .0168 \\ 
MINet$_{20}$~\cite{2020MINet}  	& CN & .9040 & .8344 & .8174 & .7705 & .9442 & .9346 & .9243 & .0093
									   & .9040 & .8761 & .8574 & .8251 & .9545 & .9454 & .9423 & .0144 \\ 
ITSD$_{20}$~\cite{2020ITSD}  		& CN & .9050 & .8523 & .8221 & .7421 & .9556 & .9407 & .9103 & .0106
									   & .9050 & .8735 & .8502 & .8068 & .9601 & .9482 & .9335 & .0165 \\ 
GateNet$_{20}$~\cite{2020GateNet} & CN & .9114 & .8566 & .8228 & .7109 & .9610 & .9385 & .8909 & .0095
									    & .9186 & .8871 & .8679 & .8229 & .9664 & .9538 & .9428 & .0137 \\ 
CSNet$_{20}$~\cite{2020CSNet} 	& CN & .8364 & .8341 & .7656 & .6319 & .9535 & .8929 & .8339 & .0169
									   & .8910 & .8790 & .8285 & .7615 & .9628 & .9171 & .9068 & .0186 \\
SAMNet$_{21}$~\cite{2021SAMNet} 	& CN & .8622 & .7813 & .7214 & .6114 & .9421 & .8700 & .8284 & .0132
									   & .8761 & .8137 & .7531 & .6843 & .9478 & .8818 & .8656 & .0217 \\	
HVPNet$_{21}$~\cite{2021HVPNet} 	& CN & .8734 & .8036 & .7377 & .6202 & .9482 & .8721 & .8270 & .0110
									   & .8610 & .7938 & .7396 & .6726 & .9320 & .8717 & .8471 & .0225 \\								   
SUCA$_{21}$~\cite{2021SUCA}  	& CN & .8988 & .8229 & .7949 & .7260 & .9520 & .9277 & .9082 & .0097
									   & .8989 & .8484 & .8237 & .7748 & .9584 & .9400 & .9194 & .0145 \\
PA-KRN$_{21}$~\cite{2021PAKRN}  & CN & .9192 & .8639 & .8358 & .7993 & .9616 & .9536 & .9416 & .0104
									   & .9239 & .8890 & .8727 & .8548 & .9680 & .9620 & .9579 & .0139 \\									   
VST$_{21}$~\cite{2021VST}        & TN & .9208 & .8716 & .8263 & .7089 & .9743 & .9442 & .8941 & .0067
									   & .9365 & .9095 & .8817 & .8262 & .9810 & .9621 & .9466 & .0094 \\
DPORTNet$_{22}$~\cite{2022DPORTNet}  &  CN & .8960 & .8363 & .7937 & .7545 & .9423 & .9116 & .9150 & .0150 
									    & .8827 & .8309 & .8184 & .7970 & .9214 & .9139 & .9083 & .0220 \\
DNTD$_{22}$~\cite{2022DNTD}  &  CN & .8957 & .8189 & .7962 & .7288 & .9378 & .9225 & .9047 & .0113 
									    & .8698 & .8231 & .8020 & .7645 & .9286 & .9086 & .9081 & .0217 \\
ICON$_{23}$~\cite{2023ICON}  &  TN & .9185 & .8622 & .8371 & .8065 & .9687 & .9619 & .9497 & .0073 
									    & .9256 & .8939 & .8671 & .8444 & .9704 & .9637 & .9554 & .0116  \\ 
\hline
LVNet$_{19}$~\cite{2019LVNet}  	  & CO & .8630 & .7794 & .7328 & .6284 & .9254 & .8801 & .8445 & .0146 
									      & .8815 & .8263 & .7995 & .7506 & .9456 & .9259 & .9195 & .0207\\
DAFNet$_{21}$~\cite{2021DAFNet}    & CO & .9166 & .8614 & .7845 & .6427 &  \textcolor{red}{\textbf{.9861}} & .9291 & .8446 & \textcolor{blue}{\textbf{.0060}}
									      & .9191 & .8928 & .8511 & .7876 & .9771 & .9539 & .9360 & .0113 \\ 
SARNet$_{21}$~\cite{2021SARNet} & CO & .9240 & .8719 & .8541 & .8304 & .9620 & .9555 & .9536 & .0099
									   & .9134 & .8850 & .8619 & .8512 & .9557 & .9477 & .9464 & .0187  \\							     
MJRBM$_{22}$~\cite{2022MJRBM} & CO & .9197 & .8656 & .8239 & .7066 & .9646 & .9350 & .8897 & .0099
									   & .9204 & .8842 & .8566 & .8022 & .9623 & .9415 & .9328 & .0163  \\
EMFINet$_{22}$~\cite{2022EMFINet} & CO & .9290 & .8720 & .8486 & .7984 & .9711 & .9604 & .9501 & .0084
									     & .9366 & .9002 & .8856 & .8617 & .9737 & .9671 & .9663 & .0109  \\
ERPNet$_{22}$~\cite{2022ERPNet}  & CO & .9210 & .8632  &.8304 & .7554  & .9603 & .9401 &.9228 & .0089 
									   & .9254  & .8974 &.8745 & .8356 & .9710 & .9566   &.9520 & .0135  \\
ACCoNet$_{22}$~\cite{2022ACCoNet} 	  & CO & .9290 & .8837 & .8552 & .7969 & .9727 & .9653 & .9450 & .0074
									   & \textcolor{blue}{\textbf{.9437}} &  .9149 & .8971 & .8806 & .9796 & .9754 & .9721 & .0088 \\
CorrNet$_{22}$~\cite{2022CorrNet} 		& CO & .9289 & .8778 & .8620 & .8311 & .9696 & .9646 & .9593 & .0083
									   &  .9380 &  .9129 & .9002 & .8875 & .9790 & .9746 & .9721 & .0098  \\	
MCCNet$_{22}$~\cite{2022MCCNet} 	  & CO & .9327 & \textcolor{blue}{\textbf{.8904}} & .8604 & .8137 & .9755 & .9685 & .9538 & .0066
				       					  & \textcolor{blue}{\textbf{.9437}} & .9155 & .9054 & .8957 & .9800 & .9758 & .9735 & .0087 \\				       					
HFANet$_{22}$~\cite{2022HFANet}  &  TO &  \textcolor{red}{\textbf{.9380}} & .8876 & \textcolor{blue}{\textbf{.8681}} & .8365 & .9740 & .9679 & .9644 & .0070 
									    & .9399 & .9112 & .8981 & .8819 & .9770 & .9712 & .9722 & .0092 \\
									   
\hline
\hline
\textbf{Ours-VGG}				  &  CO &  .9241 &  .8721 &  .8616 & .8382 & .9723 & .9636 & .9622 & .0080 
									   & .9252  & .9023 & .8932 & .8806 & .9744 & .9651  & .9655 & .0130  \\
\textbf{Ours-Res}				  & CO & .9271 & .8723  & .8621 & .8481 & .9692 & .9651 & .9644 & .0071 
									   & .9307  & .9042 & .8934 & .8826 & .9774 & .9714 & .9709 & .0098  \\
\textbf{Ours-SwinT}			  & TO & .9259 & .8774 & .8649 & \textcolor{blue}{\textbf{.8528}} & .9794 & \textcolor{blue}{\textbf{.9752}} & \textcolor{blue}{\textbf{.9713}} & \textcolor{red}{\textbf{.0055}} 
									   & .9410 & \textcolor{blue}{\textbf{.9203}} & \textcolor{blue}{\textbf{.9093}} & \textcolor{red}{\textbf{.9038}} & \textcolor{blue}{\textbf{.9829}} & \textcolor{blue}{\textbf{.9779}} & \textcolor{red}{\textbf{.9805}} & \textcolor{blue}{\textbf{.0080}}  \\

\textbf{Ours-PVT}				 & TO & \textcolor{blue}{\textbf{.9376}} & \textcolor{red}{\textbf{.8923}} & \textcolor{red}{\textbf{.8781}} & \textcolor{red}{\textbf{.8641}} & \textcolor{blue}{\textbf{.9828}} & \textcolor{red}{\textbf{.9766}} & \textcolor{red}{\textbf{.9750}} & .0064 
									   & \textcolor{red}{\textbf{.9469}} & \textcolor{red}{\textbf{.9254}} & \textcolor{red}{\textbf{.9128}} &  \textcolor{blue}{\textbf{.9035}} & \textcolor{red}{\textbf{.9860}} &  \textcolor{red}{\textbf{.9815}} &  \textcolor{blue}{\textbf{.9786}} &  \textcolor{red}{\textbf{.0079}}  \\

\toprule[1pt]
\multicolumn{18}{l}{\small{CN: CNN-based NSI-SOD method, TN: Transformer-based NSI-SOD method, CO: CNN-based ORSI-SOD method, TO: Transformer-based ORSI-SOD method.}} \\
\end{tabular}
}
\end{table*}

\begin{figure*}
\centering
\footnotesize
  \begin{overpic}[width=1\textwidth]{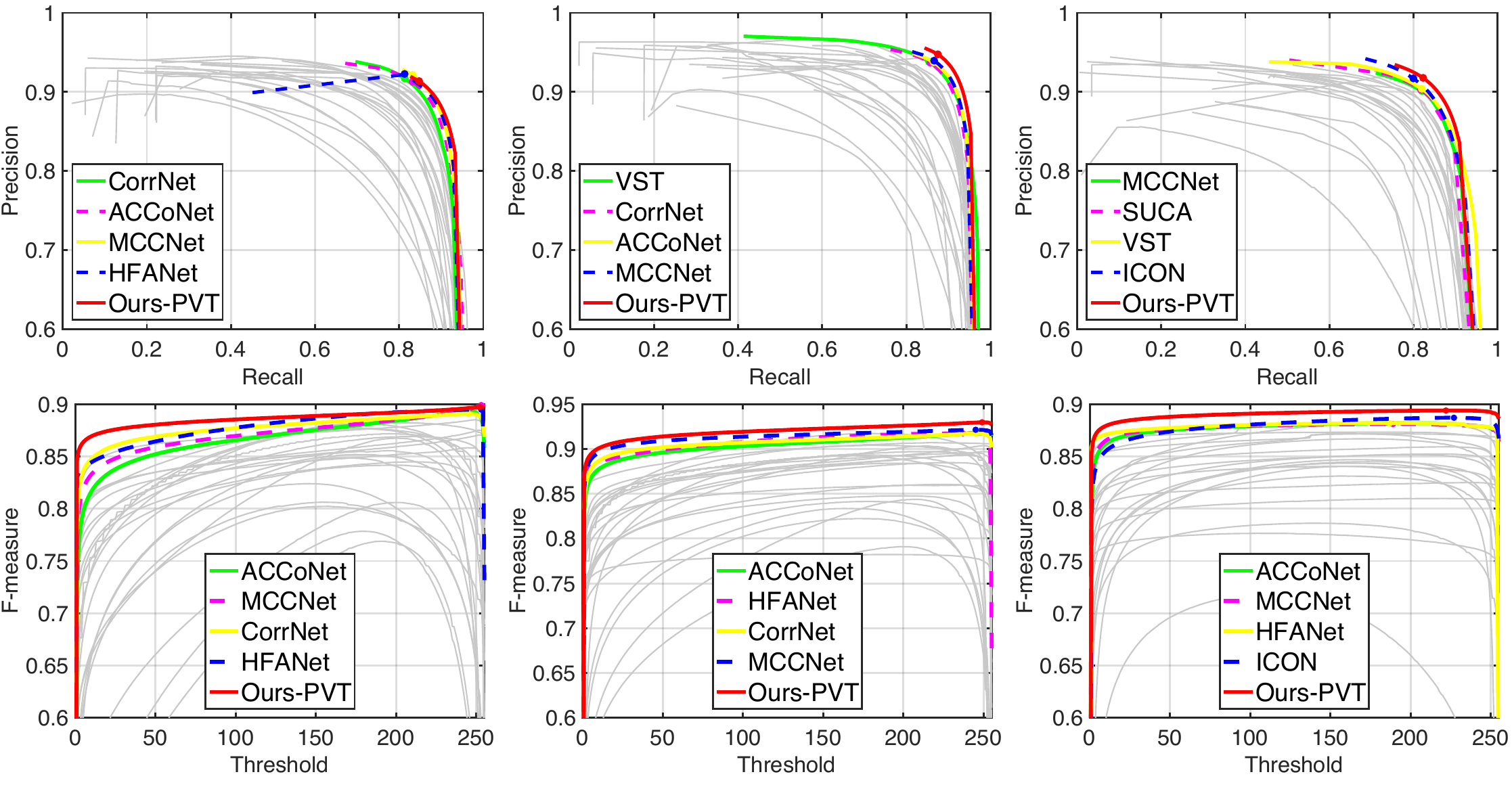}
    \put(12.6,-1.0){ (a) EORSSD\cite{2021DAFNet} }
    \put(46.7,-1.0){ (b) ORSSD\cite{2019LVNet} }   
    \put(79.3,-1.0){ (c) ORSI-4199\cite{2022MJRBM} }   
  \end{overpic}
\caption{Quantitative comparison on PR curve (the first row) and F-measure curve (the second row) in three datasets. We show the top five methods in different colors and the other compared methods in gray.}
\label{PR_Fm_comparison}
\end{figure*}

\subsection{Saliency Predictor}
\label{sec:Saliency_Predictor}
To make better use of the informative output features of D-SWSAM, KTM and SWSAM, \ie $\boldsymbol{{f}}_{\rm dswsa}$, $\boldsymbol{{f}}_{\rm ktm}$ and $\boldsymbol{{f}}_{\rm swsa}$, we adopt the effective partial decoder~\cite{2019CPD} as our saliency predictor to generate the saliency map.
Normally, the resolutions of input features in the original partial decoder are 1$\times$, 2$\times$, and 4$\times$.
However, the resolutions of input features of our saliency predictor are $32\times11\times11$ ($\boldsymbol{{f}}_{\rm swsa}$), $32\times22\times22$ ($\boldsymbol{{f}}_{\rm ktm}$), and $32\times88\times88$ ($\boldsymbol{{f}}_{\rm dswsa}$).
Therefore, we make a small modification to the original partial encoder, \ie modify the upsampling rate, to adapt to the resolutions of our input features.
In this way, our saliency predictor can generate an initial saliency map $\mathbf{s} \in [0,1]^{1\!\times\!88\!\times\!88}$.
We restore its resolution to the same resolution as the input ORSI by a 4$\times$ upsampling operation, and obtain the final saliency map $\mathbf{S} \in [0,1]^{1\!\times\!352\!\times\!352}$.

During the training phase, we train the proposed GeleNet with a hybrid loss function~\cite{21HAINet,2019BASNet}, including the intersection-over-union (IoU) loss and the binary cross-entropy (BCE) loss.
We formulate the total loss function ${L}_{\rm total}$ as follows:
\begin{equation}
   \begin{aligned}
    {L}_{\rm total}  = {\ell}_{iou} (\mathbf{S}, \mathbf{G}) + {\ell}_{bce} (\mathbf{S}, \mathbf{G}) ,
    \label{eq:SalLoss}
    \end{aligned}
\end{equation}
where ${\ell}_{iou}(\cdot)$ and ${\ell}_{bce}(\cdot)$ are IoU loss and BCE loss, respectively, and $\mathbf{G} \in \{0,1\}^{1\!\times\!352\!\times\!352}$ is the ground truth (GT). 

\section{Experiments}
\label{sec:exp}

\subsection{Experimental Setup}
\label{sec:ExpProtocol}
\textit{1) Datasets.}
We conduct experiments on the ORSSD~\cite{2019LVNet}, EORSSD~\cite{2021DAFNet}, and ORSI-4199~\cite{2022MJRBM} datasets.
The ORSSD dataset is the first public dataset for ORSI-SOD, and contains 800 images and corresponding pixel-level GTs, of which 600 images are used for training and 200 images for testing.
The EORSSD dataset contains 2,000 images and corresponding GTs, of which 1,400 images are used for training and 600 images for testing.
The ORSI-4199 dataset is the biggest dataset for ORSI-SOD, and contains 4,199 images and corresponding GTs, of which 2,000 images are used for training and 2,199 images for testing.
Following~\cite{2021DAFNet,2022EMFINet,2022CorrNet}, we train our GeleNet on the training set of each dataset and test it on the test set of each dataset.

\textit{2) Network Implementation.}
All experiments are conducted on PyTorch~\cite{PyTorch} with an NVIDIA Titan X GPU (12GB memory).
To balance the effectiveness and efficiency, we adopt PVT-v2-b2~\cite{2022PVTv2} as the backbone, and initialize it with the pre-trained parameters.
Newly added layers are all initialized with the ``Kaiming" method~\cite{InitialWei}.
We adopt rotation and a combination of flipping and rotation for data augmentation, and resize the input image and GT to 352$\times$352.
Our GeleNet is trained using Adam optimizer~\cite{Adam} for 45 epochs with a batch size of 8 and a base learning rate of $1e^{-4}$ which will decay to 1/10 every 30 epochs.

\textit{3) Evaluation Metrics.}
We adopt some widely used evaluation metrics to quantitatively evaluate the performance of our method and all compared methods on three datasets, including S-measure ($S_{\alpha}$, $\alpha$ = 0.5)~\cite{Fan2017Smeasure},
F-measure ($F_{\beta}$, $\beta^{2}$ = 0.3)~\cite{Fmeasure},
E-measure ($E_{\xi}$)~\cite{Fan2018Emeasure},
mean absolute error (MAE, $\mathcal{M}$),
precision-recall (PR) curve, and F-measure curve.
Here we adopt the evaluation tool (Matlab version)\footnote{https://github.com/MathLee/MatlabEvaluationTools} for convenient evaluation.

\begin{table}[t!]
  \centering
  \small
  \renewcommand{\arraystretch}{1.25}
  \renewcommand{\tabcolsep}{0.8mm}
  \caption{
    Quantitative comparisons with state-of-the-art NSI-SOD and ORSI-SOD methods on the ORSI-4199 dataset.
   We mark the top two results in \textcolor{red}{\textbf{red}} and \textcolor{blue}{\textbf{blue}}, respectively.
    }
\label{table:QuantitativeResults4199}
  
\resizebox{0.49\textwidth}{!}{
\begin{tabular}{r|c|cccccccc}
\midrule[1pt]    
 \multirow{2}{*}{\normalsize{Methods}}
 & \multirow{2}{*}{\normalsize{Type}}
 & \multicolumn{8}{c}{ORSI-4199~\cite{2022MJRBM}}   \\
 
 \cline{3-10}
       &  & $S_{\alpha}\uparrow$ & $F_{\beta}^{\rm{max}}\uparrow$ & $F_{\beta}^{\rm{mean}}\uparrow$ & $F_{\beta}^{\rm{adp}}\uparrow$ & $E_{\xi}^{\rm{max}}\uparrow$ & $E_{\xi}^{\rm{mean}}\uparrow$ & $E_{\xi}^{\rm{adp}}\uparrow$ & $ \mathcal{M}\downarrow$\\
	     
\midrule[1pt]
R3Net$_{18}$~\cite{2018R3Net}   	& CN & .8142 & .7847 & .7790& .7776 & .8880 & .8722 & .8645 & .0401 \\
PiCANet$_{18}$~\cite{2020PiCANet}  & CN & .7114 & .6461 & .5684 & .5933 & .7946 & .6927 & .7511 & .0974 \\ 
PoolNet$_{19}$~\cite{2019PoolNet}  & CN & .8271 & .8010 & .7779 & .7382 & .8964 & .8676 & .8531 & .0541 \\ 
EGNet$_{19}$~\cite{2019EGNet}  	& CN & .8464 & .8267 & .8041 & .7650 & .9161 & .8947 & .8620 & .0440  \\									    
BASNet$_{19}$~\cite{2019BASNet}  	& CN & .8341 & .8157 & .8042 & .7810 & .9069 & .8881 & .8882 & .0454   \\									    
CPD$_{19}$~\cite{2019CPD}  	& CN & .8476 & .8305 & .8169 & .7960 & .9168 & .9025 & .8883 & .0409   \\									    
RAS$_{20}$~\cite{2020RAS}  	& CN & .7753 & .7343 & .7141 & .7017 & .8481 & .8133 & .8308 & .0671 \\
CSNet$_{20}$~\cite{2020CSNet} 	& CN  & .8241 & .8124 & .7674 & .7162 & .9096 & .8586 & .8447 & .0524 \\
SAMNet$_{21}$~\cite{2021SAMNet} & CN & .8409 & .8249 & .8029 & .7744 & .9186 & .8938 & .8781 & .0432 \\
HVPNet$_{21}$~\cite{2021HVPNet} 	& CN & .8471 & .8295 & .8041 & .7652 & .9201 & .8956 & .8687 & .0419 \\								   
ENFNet$_{21}$~\cite{2021ENFNet}  & CN & .7766 & .7285 & .7177 & .7271 & .8370 & .8107 & .8235 & .0608 \\
SUCA$_{21}$~\cite{2021SUCA}       & CN & .8794 & .8692 & .8590 & .8415 & .9438 & .9356 & .9186 & .0304 \\
PA-KRN$_{21}$~\cite{2021PAKRN}  & CN & .8491 & .8415 & .8324 & .8200 & .9280 & .9168 & .9063 & .0382 \\
VST$_{21}$~\cite{2021VST}        & TN & .8790 & .8717 & .8524 & .7947 & .9481 & .9348 & .8997 & \textcolor{blue}{\textbf{.0281}} \\
DPORTNet$_{22}$~\cite{2022DPORTNet}  &  CN & .8094 & .7789 & .7701 & .7554 & .8759 & .8687 & .8628 & .0569 \\
DNTD$_{22}$~\cite{2022DNTD}  &  CN & .8444 & .8310 & .8208 & .8065 & .9158 & .9050 & .8963 & .0425 \\
ICON$_{23}$~\cite{2023ICON}  &  TN & .8752 & .8763 & .8664 & .8531 & .9521 & .9438 & .9239 & .0282 \\ 
\hline
MJRBM$_{22}$~\cite{2022MJRBM}   & CO & .8593 & .8493 & .8309 & .7995 & .9311 & .9102 & .8891 & .0374 \\
EMFINet$_{22}$~\cite{2022EMFINet} & CO & .8675 & .8584 & .8479 & .8186 & .9340 & .9257 & .9136 & .0330  \\
ERPNet$_{22}$~\cite{2022ERPNet}  &CO & .8670 & .8553 & .8374 & .8024 & .9290 & .9149 & .9024 & .0357 \\
ACCoNet$_{22}$~\cite{2022ACCoNet} & CO & .8775 & .8686 & .8620 & .8581 & .9412 & .9342 & .9167 & .0314 \\
CorrNet$_{22}$~\cite{2022CorrNet} 	    & CO & .8623 & .8560 &  .8513 & .8534 & .9330 & .9206 & .9142 & .0366  \\	
MCCNet$_{22}$~\cite{2022MCCNet}    & CO & .8746 & .8690 & .8630 & .8592 & .9413 & .9348 & .9182 & .0316 \\		
HFANet$_{22}$~\cite{2022HFANet} & TO & .8767 & .8700 & .8624 & .8323 & .9431 & .9336 & .9191 & .0314 \\ 

\hline
\hline
\textbf{Ours-VGG}				  & CO & .8540 & .8444 & .8374 & .8345 & .9283 & .9098 & .9086 & .0391  \\	
\textbf{Ours-Res}				  & CO & .8670 & .8601 & .8549 & .8516 & .9383 & .9284 & .9178 & .0329  \\	
\textbf{Ours-SwinT}			 & TO & \textcolor{blue}{\textbf{.8828}} & \textcolor{blue}{\textbf{.8806}} & \textcolor{blue}{\textbf{.8734}} & \textcolor{blue}{\textbf{.8681}} & \textcolor{blue}{\textbf{.9537}} & \textcolor{red}{\textbf{.9482}} & \textcolor{blue}{\textbf{.9261}} & \textcolor{red}{\textbf{.0264}} \\ 

\textbf{Ours-PVT}				 & TO & \textcolor{red}{\textbf{.8862}} & \textcolor{red}{\textbf{.8842}} & \textcolor{red}{\textbf{.8785}} & \textcolor{red}{\textbf{.8755}} & \textcolor{red}{\textbf{.9544}} & \textcolor{blue}{\textbf{.9478}} & \textcolor{red}{\textbf{.9265}} & \textcolor{red}{\textbf{.0264}}  \\									   
									   								   
\toprule[1pt]
\end{tabular}
}
\end{table}

\subsection{Comparison with State-of-the-arts}
We compare our GeleNet with state-of-the-art NSI-SOD and ORSI-SOD methods on the EORSSD and ORSSD datasets,
including R3Net~\cite{2018R3Net}, PoolNet~\cite{2019PoolNet}, EGNet~\cite{2019EGNet}, GCPA~\cite{2020GCPA}, MINet~\cite{2020MINet}, ITSD~\cite{2020ITSD}, GateNet~\cite{2020GateNet}, CSNet~\cite{2020CSNet}, SAMNet~\cite{2021SAMNet}, HVPNet~\cite{2021HVPNet}, SUCA~\cite{2021SUCA}, PA-KRN~\cite{2021PAKRN}, VST~\cite{2021VST},
DPORTNet~\cite{2022DPORTNet}, DNTD~\cite{2022DNTD}, ICON~\cite{2023ICON} with PVT backbone,
LVNet~\cite{2019LVNet}, DAFNet~\cite{2021DAFNet}, SARNet~\cite{2021SARNet}, MJRBM~\cite{2022MJRBM}, EMFINet~\cite{2022EMFINet}, ERPNet~\cite{2022ERPNet}, ACCoNet~\cite{2022ACCoNet}, CorrNet~\cite{2022CorrNet}, MCCNet~\cite{2022MCCNet}, and HFANet~\cite{2022HFANet}.
The saliency maps for the above methods are obtained from authors and public benchmarks\footnote{https://li-chongyi.github.io/proj\_optical\_saliency.html}$^{,}$\footnote{https://github.com/rmcong/DAFNet\_TIP20}~\cite{2019LVNet,2021DAFNet}, or by running public source codes.
For the ORSI-4199 dataset, we compare our GeleNet with 19 of the above 26 methods, whose saliency maps on the ORSI-4199 dataset are available, and additional five NSI-SOD methods (\ie PiCANet~\cite{2020PiCANet}, BASNet~\cite{2019BASNet}, CPD~\cite{2019CPD}, RAS~\cite{2020RAS}, ENFNet~\cite{2021ENFNet}) provided by the public benchmark\footnote{https://github.com/wchao1213/ORSI-SOD}~\cite{2022MJRBM}.
Here, for a comprehensive comparison, in addition to GeleNet with the backbone of PVT-v2-b2 (\textit{i}.\textit{e}., Ours-PVT), we also provide three variants of our GeleNet with backbones of VGG, ResNet, and Swin Transformer, named Ours-VGG, Ours-Res, and Ours-SwinT, respectively.

\begin{figure*}[t!]
    \centering
    \footnotesize
	\begin{overpic}[width=1\textwidth]{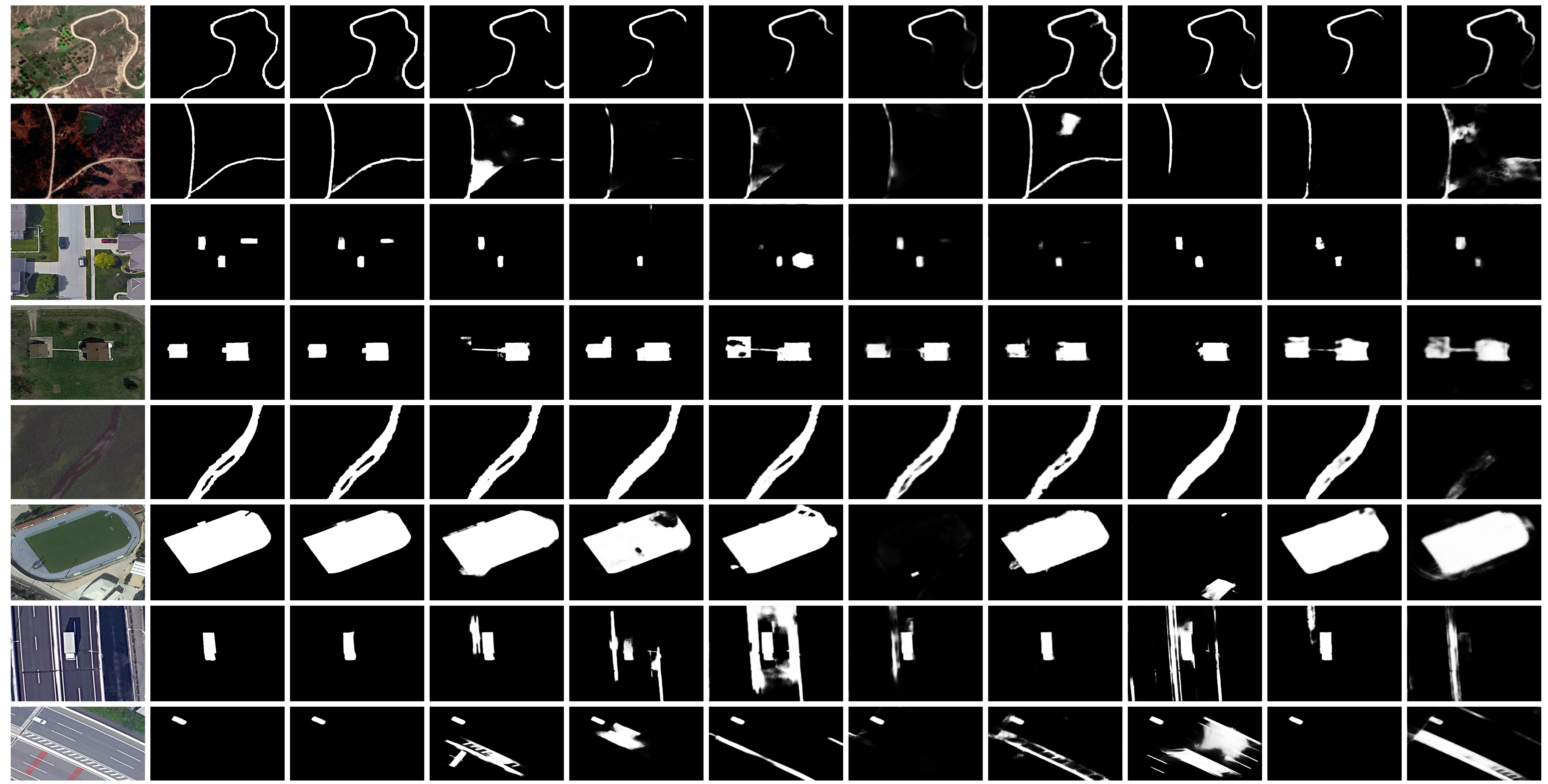}

    \put(2.75,-1.25){ ORSI }
    \put(12.5,-1.25){ GT}
    \put(18.95,-1.25){ \textbf{Ours-PVT}}
    \put(28.63,-1.25){ MCCNet  }
    \put(38.01,-1.25){ CorrNet }
    \put(46.94,-1.25){ ERPNet }
    \put(55.9,-1.25){ MJRBM }
    \put(66.05,-1.25){ VST }
    \put(73.7,-1.25){ PA-KRN }
    \put(83.6,-1.25){ SUCA }
    \put(91.87,-1.25){ HVPNet } 
    
    \end{overpic}
	\caption{Visual comparisons with eight representative state-of-the-art methods on three datasets.
    }
    \label{fig:VisualComparisons}
\end{figure*}
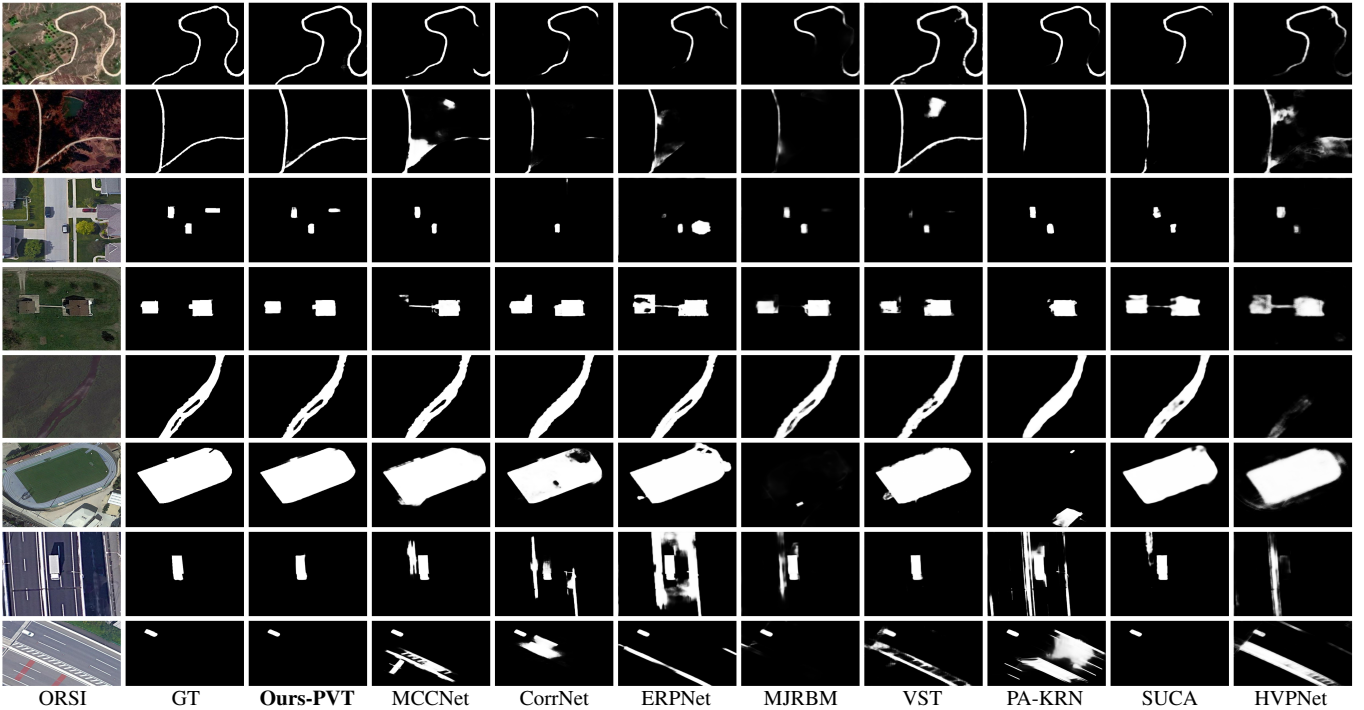

\textit{1) Quantitative Comparison on the EORSSD and ORSSD Datasets.}
We report the quantitative comparison results of our method and other 26 compared methods on the EORSSD and ORSSD datasets in Tab.~\ref{table:QuantitativeResults}.
We observe that Ours-PVT outperforms all compared methods on both datasets, except for $S_{\alpha}$, $E_{\xi}^{\rm{max}}$ and $\mathcal{M}$ on the EORSSD dataset.
Concretely, on the EORSSD dataset, Ours-PVT greatly surpasses the second-best method by 1.00\%, 2.76\%, and 1.06\% in terms of $F_{\beta}^{\rm{mean}}$, $F_{\beta}^{\rm{adp}}$, and $E_{\xi}^{\rm{adp}}$, respectively.
In $E_{\xi}^{\rm{max}}$ and $\mathcal{M}$, Ours-PVT is marginally lower than the best method by 0.33\% and 0.0004, respectively.
On the ORSSD dataset, Ours-PVT is better than the second-best method in terms of $S_{\alpha}$ (0.9469 v.s. 0.9437), $F_{\beta}^{\rm{max}}$ (0.9254 v.s. 0.9155), $E_{\xi}^{\rm{max}}$ (0.9860 v.s. 0.9810), and $\mathcal{M}$ (0.0079 v.s. 0.0087).
Notably, Ours-PVT is the only one whose $F_{\beta}^{\rm{adp}}$ exceeds 0.9, \ie 0.9035.
In addition, we plot the PR curve and F-measure curve of Ours-PVT and the compared methods on the EORSSD and ORSSD datasets in Fig.~\ref{PR_Fm_comparison} (a-b).
We can find that under different thresholds, Ours-PVT maintains its superiority and consistently achieves excellent performance.

Moreover, Ours-SwinT achieves competitive performance on the EORSSD dataset, and outperforms 26 compared methods in $F_{\beta}^{\rm{adp}}$, $E_{\xi}^{\rm{mean}}$, $E_{\xi}^{\rm{adp}}$, and $\mathcal{M}$.
Ours-SwinT ranks first out of seven metrics and second out of one metric compared to 26 compared methods on the ORSSD dataset.
Since our modules are designed specifically for the global features of transformer, the performance of our two CNN-based variants, \ie Ours-VGG and Ours-Res, is inferior to that of Ours-SwinT and Ours-PVT, and is comparable to that of ERPNet, EMFINet, and CorrNet.

\textit{2) Quantitative Comparison on the ORSI-4199 Dataset.}
Due to slight differences in the comparison methods, we report the quantitative comparison results of Ours-PVT and other 24 compared methods on the ORSI-4199 dataset separately in Tab.~\ref{table:QuantitativeResults4199}.
The ORSI-4199 dataset is the biggest and the most challenging dataset for ORSI-SOD.
The performance of Ours-PVT on this dataset is impressive, outperforming the second-best method by 0.23\%$\sim$1.63\% in terms of S-measure, F-measure, and E-measure.
And the MAE score of Ours-PVT is only 0.0264, which is one of only three methods with the MAE score below 0.03.
The advantage of Ours-PVT is easier to spot on the PR curve and F-measure curve, especially the latter one, as plotted in Fig.~\ref{PR_Fm_comparison} (c).
The above excellent performance on the challenging ORSI-4199 dataset strongly demonstrates the effectiveness of Ours-PVT.
But to be honest, there is still a lot of room for improvement on the ORSI-4199 dataset.

Ours-SwinT consistently outperforms all compared methods in all eight metrics on the ORSI-4199 dataset, and achieves the best performance in $E_{\xi}^{\rm{mean}}$ and $\mathcal{M}$, even compared to Ours-PVT.
Similar to the performance on the EORSSD and ORSSD datasets, the performance of Ours-VGG and Ours-Res is relatively average on the ORSI-4199 dataset, which further confirms that our modules is specifically designed for the global features of transformer.

In addition, Ours-PVT and two other transformer-based method (\ie VST and ICON) perform almost the best among their respective types of methods, \ie ORSI-SOD method and NSI-SOD method, on three datasets.
This means that transformer-based methods can continue to drive the development of ORSI-SOD.
The performance of the specialized ORSI-SOD method is generally better than that of NSI-SOD method on three datasets, which motivates us to develop better specialized ORSI-SOD solutions.

\textit{3) Visual Comparison.}
We show the visual comparison of Ours-PVT and eight representative state-of-the-art methods in Fig.~\ref{fig:VisualComparisons}.
There are eight cases in Fig.~\ref{fig:VisualComparisons} belonging to four typical and challenging ORSI scenes from three datasets.
The first scene is objects with various orientations, which is unique to ORSIs, as in the first two cases of Fig.~\ref{fig:VisualComparisons}.
We observe that only our method accurately highlights salient objects without including background.
In contrast, another transformer-based method, \ie VST, incorrectly highlights some background regions, and all CNN-based methods fail to fully highlight objects.
This is attributed to the directional convolution unit of D-SWSAM.
The second scene contains multiple salient objects, as in the third and fourth cases of Fig.~\ref{fig:VisualComparisons}.
Most methods only locate some of these objects and their saliency maps are relatively rough, but our method finely segments all salient objects.
This is due to the precise location capability of SWSAM.
The third scene contains objects with fine structure, as in the fifth and sixth cases of Fig.~\ref{fig:VisualComparisons}.
Our method successfully delineates the same fine structure of salient objects as GTs, such as the islands in the river and the shape of the playground.
The last scene is low contrast, where the color of foreground and background is similar, as in the last two cases of Fig.~\ref{fig:VisualComparisons}.
Due to the global modeling capability of PVT and the local enhancement of proposed modules, our method accurately distinguishes white vehicles in both cases without the interference of white zebra crossings.
While other methods are confused by the white zebra crossing and wrongly highlight them.

\subsection{Ablation Studies}
\label{Ablation Studies}
We conduct comprehensive ablation studies on the EORSSD dataset to evaluate the effectiveness of each module of our GeleNet and each component of our three modules.
Accordingly, we analyze
1) the individual contribution of three modules,
2) the effectiveness of each component in D-SWSAM,
3) the rationality of the way of modeling knowledge in KTM, and
4) the effectiveness of each component in SWSAM.
We conduct these ablation studies on the GeleNet with the backbone of PVT-v2-b2, and adopt the same parameter settings and dataset partitioning as in Sec.~\ref{sec:ExpProtocol} for all variants.

\textit{1) Individual Contribution of Three Modules}.
To investigate the individual contribution of the proposed three modules, \ie D-SWSAM, KTM, and SWSAM, we design various combinations of the above three modules for a total of seven variants:
1) Baseline, in which we remove all proposed modules and adopt element-wise summation to fuse $\boldsymbol{f}^{2}_{\rm t}$ and $\boldsymbol{f}^{3}_{\rm t}$,
2) Baseline+D-SWSAM,
3) Baseline+KTM,
4) Baseline+SWSAM,
5) Baseline+KTM+SWSAM,
6) Baseline+D-SWSAM+SWSAM, and
7) Baseline+D-SWSAM+KTM.
We report the quantitative results in Tab.~\ref{Ablation_module}.

\begin{table}[!t]
\centering
\caption{Ablation results of evaluating the individual contribution of each module in GeleNet.
  The best one in each column is \textbf{bold}.
  }
\label{Ablation_module}
\renewcommand{\arraystretch}{1.25}
\renewcommand{\tabcolsep}{1.1mm}
\begin{tabular}{c|cccc||ccc}
\bottomrule

  \multirow{2}{*}{No.} & \multirow{2}{*}{Baseline} & \multirow{2}{*}{D-SWSAM} & \multirow{2}{*}{KTM} & \multirow{2}{*}{SWSAM}  
 & \multicolumn{3}{c}{EORSSD~\cite{2021DAFNet}}  \\
 
 \cline{6-8}
    & & & & 
    &$S_{\alpha}\uparrow$ & $F_{\beta}^{\rm{max}}\uparrow$ 
    & $E_{\xi}^{\rm{max}}\uparrow$   \\
\hline
\hline
1 &  \Checkmark &                      &                      &                      & 0.9249 & 0.8717 & 0.9712   \\
2 &  \Checkmark & \Checkmark  &                      &                      & 0.9305 & 0.8827 & 0.9764   \\
3 &  \Checkmark &                      & \Checkmark &                       & 0.9301 & 0.8812 & 0.9778  \\
4 &  \Checkmark &                      &                     & \Checkmark  & 0.9309 & 0.8836  & 0.9775   \\
\hline
5 &  \Checkmark &                      & \Checkmark & \Checkmark  & 0.9350 & 0.8872 & 0.9796   \\
6 &  \Checkmark & \Checkmark  &                     & \Checkmark  & 0.9328 & 0.8871 & 0.9786   \\ 
7 &  \Checkmark & \Checkmark  & \Checkmark &                      & 0.9339 & 0.8863 & 0.9791   \\ 
\hline
8 &  \Checkmark & \Checkmark$^{*}$ & \Checkmark &                      & 0.9355 & 0.8879 & 0.9798   \\  
9 &  \Checkmark &                      & \Checkmark & \Checkmark$^{*}$ & 0.9366 & 0.8911 & 0.9802   \\   

\hline
10 &  \Checkmark & \Checkmark  & \Checkmark & \Checkmark  & \textbf{0.9376} & \textbf{0.8923}  & \textbf{0.9828}  \\
\toprule
\multicolumn{8}{l}{\Checkmark$^{*}$: using this module to enhance the lowest- and highest-level features.}\\
\end{tabular}
\end{table}

\begin{figure}[t!]
  \centering
  \footnotesize
  \begin{overpic}[width=1\columnwidth]{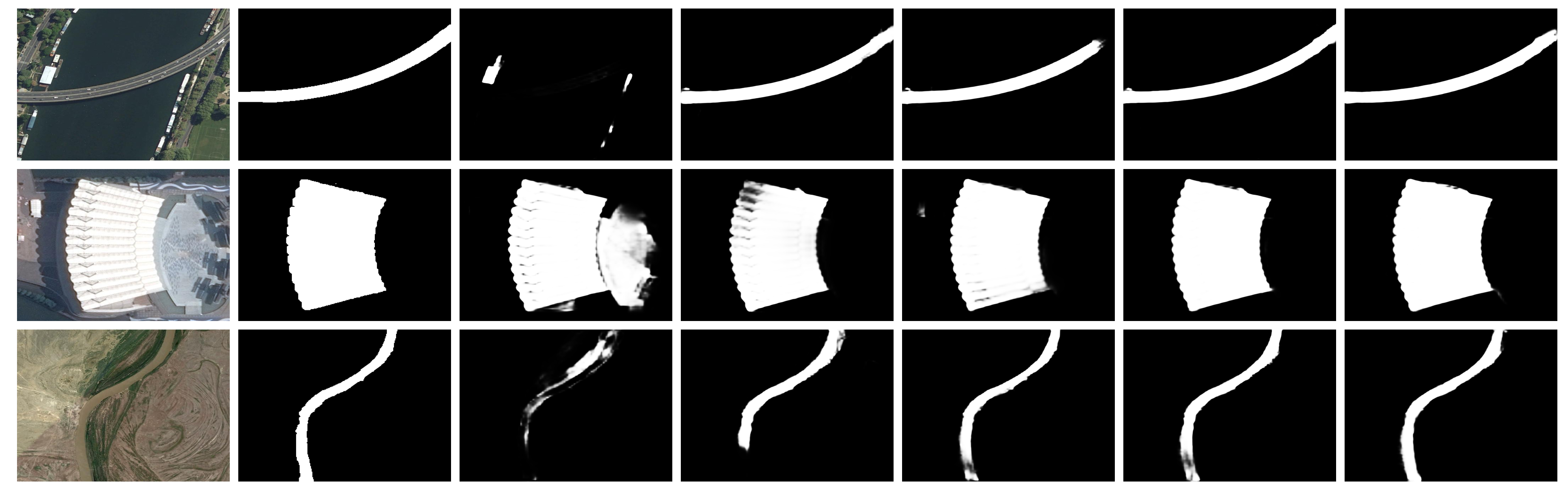}
      \put(3.,-1.7){ ORSI}
      \put(18.95,-1.7){ GT }   
      \put(32.5,-1.7){B (1)}
      \put(43.5,-1.7){ B+D (2)}      
      \put(57.9,-1.7){ B+S (4)}  
      \put(70.1,-1.7){ B+D+S (6)}
      \put(84.8,-1.7){  \textbf{Ours} (10)}
  \end{overpic}
  \caption{Visual comparisons of different variants.
  B, D, and S are Baseline, D-SWSAM, and SWSAM, respectively.
  The numbers in parentheses are the ordinal numbers of these variants in Tab.~\ref{Ablation_module}.
    }\label{fig:example_AB1}
\end{figure}
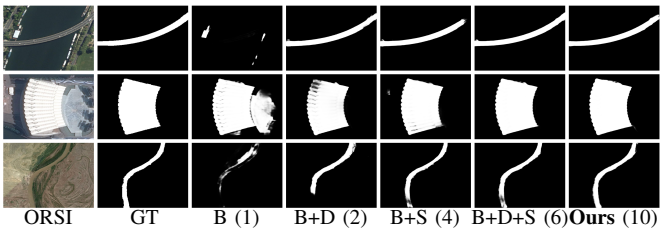

From the first four rows in Tab.~\ref{Ablation_module}, we can find that each module can individually improve ``Baseline" by around 0.5\% in $S_{\alpha}$, around 1.0\% in $F_{\beta}^{\rm{max}}$, and around 0.6\% in $E_{\xi}^{\rm{max}}$, which directly proves that the proposed three modules are effective.
The fifth to seventh rows of Tab.~\ref{Ablation_module} present the performance of pairwise cooperation of modules.
We can conclude that the cooperation of different modules can further improve the robustness of our method, resulting in better performance.
Therefore, with all three modules working together, our full model significantly outperforms ``Baseline" by 1.27\% in $S_{\alpha}$, 2.06\% in $F_{\beta}^{\rm{max}}$, and 1.16\% in $E_{\xi}^{\rm{max}}$.

We provide two variants to prove the necessity of enhancing the lowest-level and highest-level features with different modules:
8) using D-SWSAM to enhance both lowest-level and highest-level features, and
9) using SWSAM to enhance both lowest-level and highest-level features.
As shown in Tab.~\ref{Ablation_module}, we observe that the performance of the above two variants is not as good as our method with different enhancements.
This means that enhancing the lowest-level and highest-level features with the same module is suboptimal, and our different enhancements to the lowest-level and highest-level features are necessary.

\begin{table}[!t]
\centering
\caption{Ablation results of evaluating the effectiveness of each component of the proposed three modules.
  The best one in each column is \textbf{bold}. D-SW. means D-SWSAM, and SWSA. means SWSAM.
  }
\label{Ablation_component}
\renewcommand{\arraystretch}{1.35}
\renewcommand{\tabcolsep}{3.2mm}
\begin{tabular}{l|c||ccc}
\bottomrule
  & \multirow{2}{*}{Models}  & \multicolumn{3}{c}{EORSSD~\cite{2021DAFNet}}  \\
 \cline{3-5}
  &  & $S_{\alpha}\uparrow$ & $F_{\beta}^{\rm{max}}\uparrow$ 
    & $E_{\xi}^{\rm{max}}\uparrow$  \\
\hline
\hline
& GeleNet (\textbf{Ours}) & \textbf{0.9376} & \textbf{0.8923}  & \textbf{0.9828}  \\

\hline
  \multirow{2}{*}{\begin{sideways}D-SW.\end{sideways}}
& \textit{w/o DirConv}       & 0.9366 & 0.8911 & 0.9802    \\%
& \textit{w/o SWSAM}                   & 0.9346 & 0.8886 & 0.9800    \\

\hline
  \multirow{2}{*}{\begin{sideways}KTM\end{sideways}}
& \textit{w/ sum}             & 0.9353 & 0.8886 & 0.9808     \\
& \textit{w/ product}              & 0.9334 & 0.8876 & 0.9813    \\

\hline
  \multirow{2}{*}{\begin{sideways}SWSA.\end{sideways}}
& \textit{w/o shuffle}              & 0.9320 & 0.8895 & 0.9798     \\
& \textit{w/o weights}          & 0.9319 & 0.8872 & 0.9788     \\
\toprule
\end{tabular}
\end{table}

Furthermore, we show the saliency maps for the first, second, fourth, sixth variants, and our full model in Fig.~\ref{fig:example_AB1} to visually illustrate the role of modules.
Without the help of any modules, ``Baseline" performs badly, and its saliency maps have the problems of wrongly highlighting, introducing background, and incomplete highlighting.
With the addition of D-SWSAM which can perceive the orientation information and perform local enhancement, the saliency maps generated by ``B+D" successfully highlight the salient objects with the correct direction (\ie the first and last cases) and suppress the background (\ie the second case).
Since SWSAM is responsible for location enhancement in the highest-level features, the salient objects in the saliency maps generated by ``B+S" are highlighted correctly and completely.
Naturally, the combination of D-SWSAM and SWSAM, \ie ``B+D+S", inherits all the advantages of the two modules.
With the additional help of KTM, the saliency maps generated by our full model are visually indistinguishable from GTs.
The above analysis proves that the proposed three modules are effective and play their respective functions.

\textit{2) Effectiveness of Each Component in D-SWSAM.}
D-SWSAM consists of a directional convolution unit and SWSAM.
Here, we provide two variants of D-SWSAM to investigate the effectiveness of the above components:
1) removing directional convolution unit (\ie \textit{w/o DirConv} which is the same as No.9 in Tab.~\ref{Ablation_module}), and
2) removing SWSAM (\ie \textit{w/o SWSAM}).
As shown in the second and third rows of Tab.~\ref{Ablation_component}, removing either component reduces detection accuracy, which demonstrates both components are necessary for D-SWSAM.
Notably, the performance of \textit{w/o SWSAM} degrades more than that of \textit{w/o DirConv}, indicating that SWSAM  is more important in D-SWSAM.

\textit{3) Rationality of the Way of Modeling Knowledge in KTM.}
Due to the product and sum of $\boldsymbol{f}^{2}_{\rm t}$ and $\boldsymbol{f}^{3}_{\rm t}$ are complementary, we model the contextual correlation knowledge between the product and sum of $\boldsymbol{f}^{2}_{\rm t}$ and $\boldsymbol{f}^{3}_{\rm t}$ in KTM.
To investigate the rationality of this way of modeling knowledge, we design two alternative modeling strategies:
1) removing product then modeling knowledge only from sum  (\ie \textit{w/ sum}), and
2) removing sum then modeling knowledge only from product  (\ie \textit{w/ product}).
As shown in the fourth and fifth rows of Tab.~\ref{Ablation_component}, \textit{w/ sum} and \textit{w/ product} perform worse.
As detailed in Sec.~\ref{sec:KTM}, due to the complementarity between the product and sum of $\boldsymbol{f}^{2}_{\rm t}$ and $\boldsymbol{f}^{3}_{\rm t}$, the contextual correlation knowledge modeled from both is more conducive to inferring salient objects.
Modeling knowledge from only one of them is suboptimal.

\begin{table}[!t]
\centering
\caption{
Comparing the proposed SWSA with two classic attention operations, \ie the traditional spatial attention~\cite{2018CBAM} and SGE~\cite{2019SGE}.
  The best one in each column is \textbf{bold}.
  }
\label{Ablation_SA}
\renewcommand{\arraystretch}{1.25}
\renewcommand{\tabcolsep}{3.2mm}
\begin{tabular}{c||ccc}
\bottomrule
  \multirow{2}{*}{Models}  & \multicolumn{3}{c}{EORSSD~\cite{2021DAFNet}}  \\
 \cline{2-4}
   & $S_{\alpha}\uparrow$ & $F_{\beta}^{\rm{max}}\uparrow$ 
    & $E_{\xi}^{\rm{max}}\uparrow$  \\
\hline
\hline
  \textit{w/ SWSAM} (\textbf{Ours}) & \textbf{0.9376} & \textbf{0.8923}  & \textbf{0.9828}  \\

\hline

 \textit{w/ SA}           & 0.9293 & 0.8850 & 0.9784     \\
 \textit{w/ SGE}          	    & 0.9324 & 0.8843 & 0.9791   \\

\toprule
\end{tabular}
\end{table}

\textit{4) Effectiveness of Each Component in SWSAM.}
SWSAM plays an important role in our GeleNet.
We use it twice in our GeleNet on the lowest-level and highest-level features.
Here, we provide two variants of SWSAM to investigate the effectiveness of its components:
1) removing channel shuffle (\ie \textit{w/o shuffle}), and
2) removing learnable parameter ${w}^{n}$ in Eq.~\ref{eq:3} (\ie \textit{w/o weights}).
Notably, these two variants are applied in SWSAM of D-SWSAM and SWSAM of the highest level.
As shown in the last two rows of Tab.~\ref{Ablation_component}, \textit{w/o shuffle} and \textit{w/o weights} are almost the worst of all variants in Tab.~\ref{Ablation_component}, which illustrates the importance of both operations.
Actually, channel shuffle in SWSAM serves two different purposes.
\textit{w/o shuffle} lets SWSAM in D-SWSAM to generate four spatial attention maps directly from four sub-features with single direction instead of sub-features with uniform four directions, and weakens the channel communication of the global highest-level features.
\textit{w/o weights} does not take into account the differences between different spatial attention maps and simply fuses spatial attention maps.
Therefore, the performance of both variants is degraded.

\begin{figure}[t!]
  \centering
  \footnotesize
  \begin{overpic}[width=1\columnwidth]{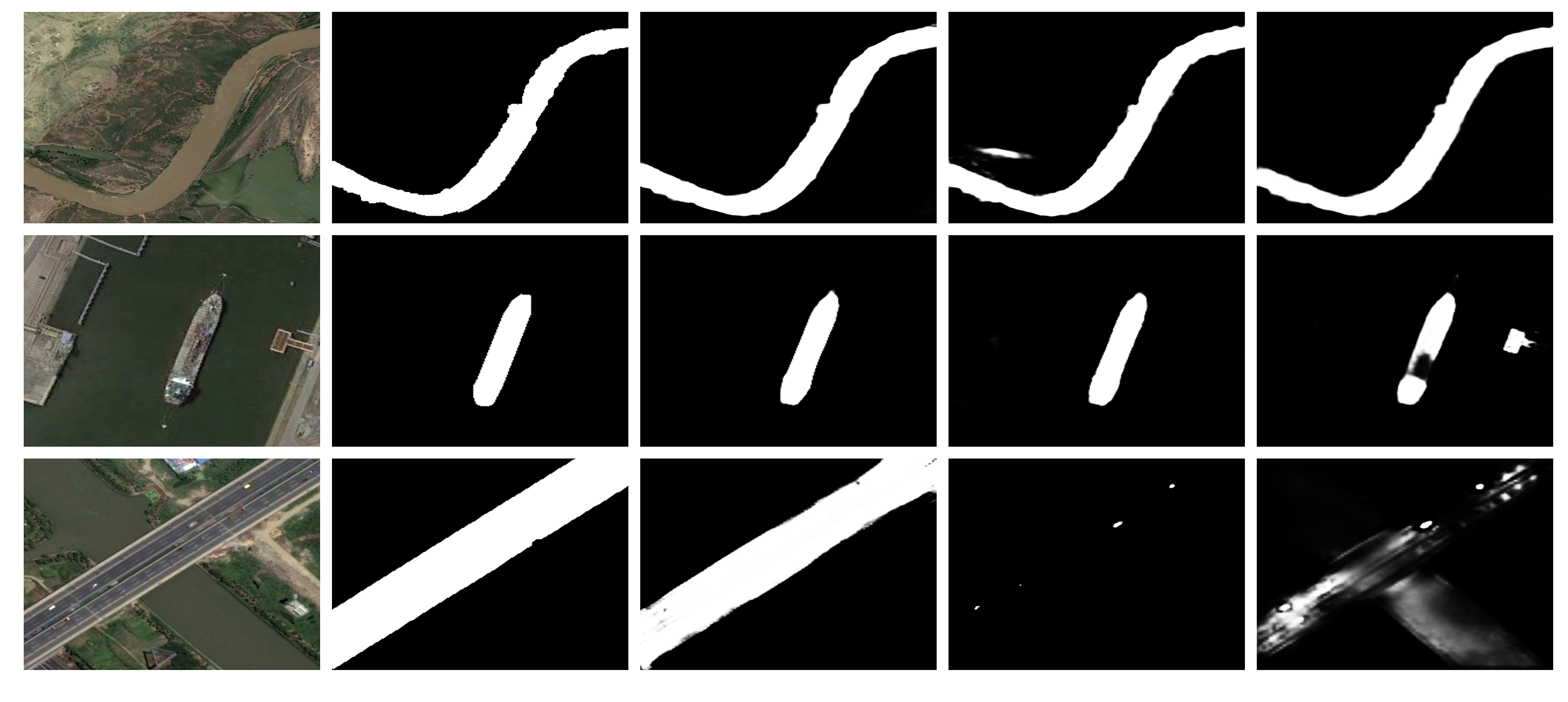}
      \put(6.,-1){ ORSI}
      \put(27.1,-1){ GT }   
      \put(36.6,-1){ \textit{w/ SWSAM} (\textbf{Ours})}      
      \put(64.9,-1){ \textit{w/ SA}}
      \put(82.85,-1){ \textit{w/ SGE} }
  \end{overpic}
  \caption{Visual comparisons of different attention mechanisms.
    }\label{fig:example_AB2}
\end{figure}
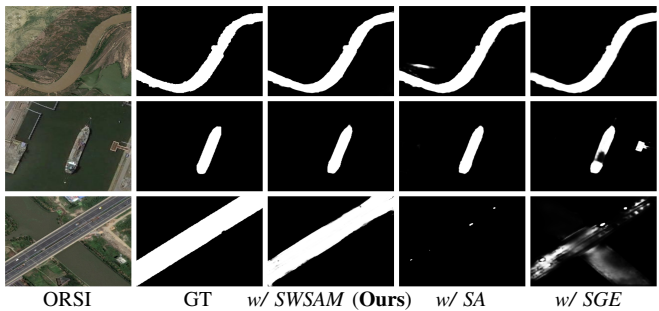

In addition, we compare the proposed SWSAM with two classic attention mechanisms, \ie the traditional spatial attention~\cite{2018CBAM} and SGE, to further investigate the effectiveness of our SWSAM.
We provide two variants:
1) replacing SWSAM with the traditional spatial attention (\ie \textit{w/ SA}), and
2) replacing SWSAM with SGE (\ie \textit{w/ SGE}).
As reported in Tab.~\ref{Ablation_SA}, the effectiveness of these two attention mechanisms is lower than that of our SWSAM, \ie \textit{w/ SWSAM}, for ORSI-SOD.
Moreover, in Fig.~\ref{fig:example_AB2}, we show the saliency maps generated by \textit{w/ SWSAM}, \textit{w/ SA}, and \textit{w/ SGE} for the visual comparison.
The first case is that some background regions are similar to salient objects.
Traditional spatial attention generates the attention map in a global manner (\ie from all channels), which leads to the omission of valid information and is not conducive to generating an accurate attention map.
Therefore, \textit{w/ SA} incorrectly highlights background regions similar to salient objects.
The second case is the scene with the irrelevant object.
Since SGE extracts specific semantics from each sub-feature and does not adopt the same consistent attention map for enhancement, \textit{w/ SGE} mistakenly highlights the irrelevant object in the scene.
The last case is the elevated highway with cars.
Since the comprehensive valid information in traditional spatial attention is omitted, \textit{w/ SA} only highlights cars on the elevated highway instead of the elevated highway.
\textit{w/ SGE} takes into account the semantics of different sub-features, so it highlights more regions than \textit{w/ SA}, but meanwhile introduces other background regions.
Differently, our SWSAM aggregates multiple attention maps generated from different sub-features in an adaptive way, resulting in a comprehensive attention map for consistent enhancement.
Therefore, our \textit{w/ SWSAM} can effectively handle the above cases.

\section{Conclusion}
\label{sec:con}
In this paper, we propose the first transformer-driven ORSI-SOD solution, namely GeleNet.
GeleNet mainly follows the global-to-local paradigm, while also considering cross-level contextual interactions.
GeleNet employs PVT to extract global features, SWSAM and D-SWSAM to achieve local enhancement, and KTM to activate cross-level contextual interactions.
Specifically, SWSAM is an improved spatial attention module, which is responsible for location enhancement in the highest-level features.
To adapt to various object orientations in ORSIs, directional convolutions are used in D-SWSAM to explicitly perceive orientation information of the lowest-level features, followed by SWSAM to achieve detail enhancement.
KTM is built on the self-attention mechanism, and models the complementary information between the product and the sum of two middle-level features to generate discriminative features.
The cooperation of components makes GeleNet a successful salient object detector for ORSIs.
Extensive comparisons and ablation studies demonstrate the superiority of GeleNet and the effectiveness of the three proposed modules.
Moreover, the proposed D-SWSAM and SWSAM can be used as plug-and-play modules for related tasks~\cite{19CRMCO,2015SODBenchmark,Tu1,VT821,2021ENFNet}.



\ifCLASSOPTIONcaptionsoff
  \newpage
\fi

\bibliographystyle{IEEEtran}
\bibliography{PVTRef}

%



%

\end{document}